\documentclass[preprint,12pt]{elsarticle}

\usepackage{lineno,hyperref}
\modulolinenumbers[5]
\usepackage{changepage}
\usepackage{amsmath}
\usepackage{graphicx}
\usepackage{subfig}
\usepackage{multirow}
\usepackage{array}
\newcolumntype{P}[1]{>{\centering\arraybackslash}p{#1}}
\usepackage{makecell}
\usepackage{pgfplots}
\usepackage{dingbat}
\usepackage[flushleft]{threeparttable}
\usepackage{alphalph}
\usepackage{bm}

\journal{ISPRS J. Photo. Remote Sens.}
\begin{document}
\begin{frontmatter}
\title{Recurrently Exploring Class-wise Attention in A Hybrid Convolutional and Bidirectional LSTM Network for Multi-label Aerial Image Classification}

\author[firstaddress,secondaryaddress]{Yuansheng Hua\corref{sameauthor}}
\cortext[sameauthor]{The first two authors contributed equally to this work.}
\ead{yuansheng.hua@tum.de}

\author[firstaddress,secondaryaddress]{Lichao Mou\corref{sameauthor}}
\ead{lichao.mou@dlr.de}

\author[firstaddress,secondaryaddress]{Xiao Xiang Zhu\corref{correspondingauthor}}
\cortext[correspondingauthor]{Corresponding author}
\ead{Xiaoxiang.Zhu@dlr.de}

\address[firstaddress]{Remote Sensing Technology Institute (IMF), German Aerospace Center (DLR),  Oberpfaffenhofen, 82234 Wessling, Germany}
\address[secondaryaddress]{Signal Processing in Earth Observation (SiPEO), Technical University of Munich (TUM),  Arcisstr. 21, 80333 Munich, Germany}

\begin{abstract}
\textbf{This is a preprint. To read the final version please visit e ISPRS Journal of Photogrammetry and Remote Sensing.} Aerial image classification is of great significance in the remote sensing community, and many researches have been conducted over the past few years. Among these studies, most of them focus on categorizing an image into one semantic label, while in the real world, an aerial image is often associated with multiple labels, e.g., multiple object-level labels in our case. Besides, a comprehensive picture of present objects in a given high-resolution aerial image can provide a more in-depth understanding of the studied region. For these reasons, aerial image multi-label classification has been attracting increasing attention. However, one common limitation shared by existing methods in the community is that the co-occurrence relationship of various classes, so-called class dependency, is underexplored and leads to an inconsiderate decision. In this paper, we propose a novel end-to-end network, namely class-wise attention-based convolutional and bidirectional LSTM network (CA-Conv-BiLSTM), for this task. The proposed network consists of three indispensable components: 1) a feature extraction module, 2) a class attention learning layer, and 3) a bidirectional LSTM-based sub-network. Particularly, the feature extraction module is designed for extracting fine-grained semantic feature maps, while the class attention learning layer aims at capturing discriminative class-specific features. As the most important part, the bidirectional LSTM-based sub-network models the underlying class dependency in both directions and produce structured multiple object labels. Experimental results on UCM multi-label dataset and DFC15 multi-label dataset validate the effectiveness of our model quantitatively and qualitatively.
\end{abstract}


\end{frontmatter}

\section{Introduction} 

\begin{figure*}[!t]
\centering
\subfloat[]{\includegraphics[width=1.35in]{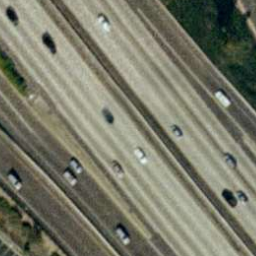}
\label{fig:ss1}}
\hspace{0.1in}
\subfloat[]{\includegraphics[width=1.35in]{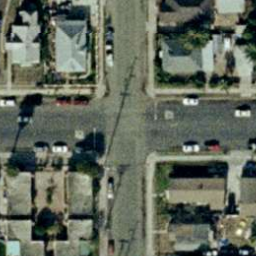}
\label{fig:ss2}}
\hspace{0.1in}
\subfloat[]{\includegraphics[width=1.35in]{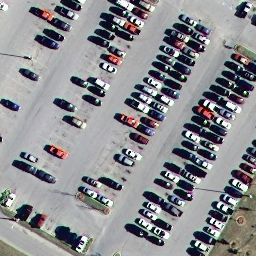}
\label{fig:ss3}}
\caption{Example high resolution aerial images with their scene labels and multiple \textit{object} labels. Common label pairs are \textbf{\textit{highlighted}}. (a) Free way: \textsl{bare soil}, \textbf{\textsl{car}}, \textsl{grass}, \textbf{\textsl{pavement}} and \textsl{tree}. (b) Intersection: \textsl{building}, \textbf{\textsl{car}}, \textsl{grass}, \textbf{\textsl{pavement}} and \textsl{tree}. (c) Parking lot: \textbf{\textsl{car}} and \textbf{\textsl{pavement}}.}
\label{fig:ss01}
\end{figure*}

With the booming of remote sensing techniques in the recent years, a huge volume of high resolution aerial imagery is now accessible and benefits a wide range of real-world applications, such as urban mapping \cite{Marmanis17, beyongrgb, Marcos18, mou2018rifcn}, ecological monitoring \cite{zarco2014tree, wen2017semantic}, geomorphological analysis \cite{Mou18, lucchesi2013applications, weng2018land, cheng2017remote}, and traffic management \cite{mou2018vehicle, 7729468, hsf}. As a fundamental bridge between aerial images and these applications, image classification, which aims at categorizing images into semantic classes, has obtained wide attention, and many researches have been conducted recently \cite{nogueira2017towards,yang2010bag, xia2017aid, zhu2017deep, 7358126, rs71114680, hu2018recent, zhang2015saliency, HUANG2018127, mou2017multitemporal}. However, most existing studies assume that each image belongs to only one label (e.g., scene-level labels in Fig. \ref{fig:ss01}), while in reality, an image is usually associated with multiple labels \cite{tan2017multi}. Furthermore, a comprehensive picture of objects present in an aerial image is capable of offering a holistic understanding of such image. With this intention, numerous researches, i.e., semantic segmentation \cite{ren2015faster, long2015fully, badrinarayanan2015segnet} and object detection \cite{ren2015faster, viola2001rapid, lin2017feature, ren2017object}, have emerged recently. Unfortunately, it is extremely labor- and time-consuming to acquire ground truths for these studies (i.e., pixel-wise segmentation masks and bounding-box-level annotations). Compared to these expensive labels, image-level labels (cf. multiple object-level labels in Fig. \ref{fig:ss01}) are at a fair low cost and readily accessible. To this end, multi-label classification, aiming at assigning an image with multiple object labels, is arising in both remote sensing \cite{karalas2016land, zeggada2017deep, koda2018spatial,zeggada2018multilabel} and computer vision communities \cite{Patterson2012SunAttributes, nus-wide-civr09, everingham2010pascal}. In this paper, we deploy our efforts in exploring an efficient multi-label classification model.

\subsection{The Challenges of Multi-label Classification}
Benefited from the fast-growing remote sensing technology, large quantities of high-resolution aerial images are available and widely used in many visual tasks. Along with such huge opportunities, challenges have come up inevitably. 

On one hand, it is difficult to extract high-level features from high-resolution images. Considering its complex spatial structure, conventional hand-crafted features, and mid-level semantic models \cite{yang2010bag, shao2013hierarchical, risojevic2013fusion, lowe2004distinctive, 7466064} suffer from the poor performance of capturing holistic semantic features, which leads to an unsatisfactory classification ability.

On the other hand, underlying correlations between dependent labels are required to be unearthed for an efficient prediction of multiple object labels. E.g., the existence of ships infers to a high probable co-occurrence of the sea, while the presence of buildings is almost always accompanied by the coexistence of pavement. However, the recently proposed multi-label classification methods \cite{karalas2016land, zeggada2017deep, koda2018spatial, zeggada2018multilabel} assumed that classes are independent and employed a set of binary classifiers \cite{karalas2016land} or a regression model \cite{zeggada2017deep, koda2018spatial, zeggada2018multilabel} to infer the existence of each class separately.

To summarize, a well-performed multi-label classification system requires powerful capabilities of learning holistic feature representations and should be capable of harnessing the implicit class dependency.

\subsection{The Motivation of Our Work}

As our survey of related work shows above, recent approaches make few efforts to exploit the high-order class dependency, which constrains the performance in multi-label classification. Besides, direct utilization of CNNs pre-trained on natural image datasets \cite{zeggada2017deep, koda2018spatial, zeggada2018multilabel} leads to a partial interpretation of aerial images due to their diverse visual patterns. Moreover, most state-of-the-art methods decompose multi-label classification into separate stages, which cuts off their inter-correlations and makes end-to-end training infeasible.

To tackle these problems, in this paper, we propose a novel end-to-end network architecture, class attention-based convolutional and bidirectional LSTM network (CA-Conv-BiLSTM), which integrates feature extraction and high-order class dependency exploitation together for multi-label classification. Contributions of our work to the literature are detailed as follows:

\begin{itemize}
  \item We regard the multi-label classification of aerial images as a structured output problem instead of a simple regression problem. In this manner, labels are predicted in an ordered procedure, and the prediction of each label is dependent on others. As a consequence, the implicit class relevance is taken into consideration, and structured outputs are more reasonable and closer to the real-world case as compared to regression outputs.
  \item we propose an end-to-end trainable network architecture for multi-label classification, which consists of a feature extraction module (e.g., a modified network based on VGG-16), a class attention learning layer, and a bidirectional LSTM-based sub-network. These components are designed for extracting features from input images, learning discriminative class-specific features, and exploiting class dependencies, respectively. Besides, such a design makes it feasible to train the network in an end-to-end fashion, which enhances the compactness of our model.
  \item Considering that class dependencies are diverse in both directions, a bidirectional analysis is required for modeling such correlations. Therefore, we employ a bidirectional LSTM-based network, instead of a one-way recurrent neural network, to dig out class relationships.
  \item We build a new challenging dataset, DFC15 multi-label dataset, by reproducing from a semantic segmentation dataset, GRSS\_DFC\_2015 (DFC15) \cite{dfc15}. The proposed dataset consists of aerial images at a spatial resolution of 5 cm and can be used to evaluate the performance of networks for multi-label classification.
  
\end{itemize}

The following sections further introduce and discuss our network. Specifically, Section \ref{sec:method} provides an intuitive illustration of the class dependency and then details the structure of the proposed network in terms of its three fundamental components. Section \ref{sec:experiment} describes the setup of our experiments, and experimental results are discussed from quantitative and qualitative perspectives. Finally, the conclusion of this paper is drawn in Section \ref{sec:conclusion}.

\section{Methodology}
\label{sec:method}
\subsection{An Observation}
\label{sec:ob}

Current aerial image multi-label classification methods  \cite{zeggada2017deep, koda2018spatial,zeggada2018multilabel} consider such problem as a regression issue, where models are trained to fit a binary sequence, and each digit indicates the existence of its corresponding class. Unlike one-hot vectors, a binary sequence is allowed to contain more than one 'hot' value for indicating the joint existence of multiple candidate classes in one image. Besides, several researches \cite{karalas2016land} formulate multi-label classification into several single-label classification tasks, and thus, train a set of binary classifiers for each class. Notably, one common assumption of these studies is that classes are independent of each other, and classifiers predict the existence of each category independently. 
\begin{figure*}[!t]
\begin{adjustwidth}{-0.6cm}{0cm}
\centering
\includegraphics[width=0.95\textwidth]{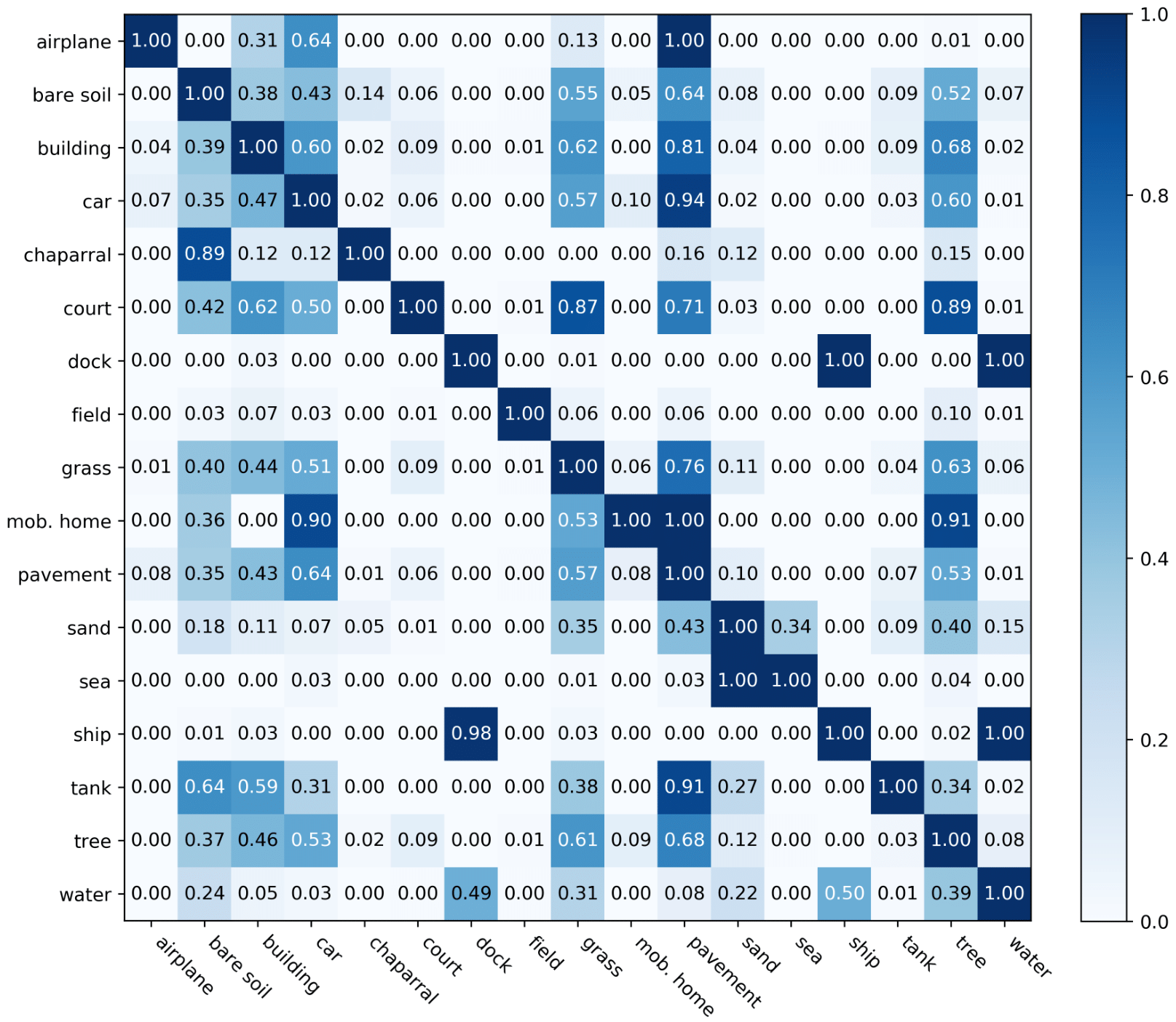}
\end{adjustwidth}
\caption{The co-occurrence matrix of labels in UCM multi-label dataset. Notably, all images are taken into consideration when calculating this matrix. Labels at Y-axis represent referenced classes $C_r$ , while labels at X-axis are potential co-occurrence classes $C_p$. The conditional probability $P(C_p|C_r)$ of each class pair is presented in the corresponding block.}
\label{fig:cm_ucm}
\end{figure*}
However, this is violent and not accord with real life. As illustrated in Fig. \ref{fig:ss01}, although images obtained in diverse scenes are assigned with multiple different labels, there are still common classes, e.g., car and pavement, coexisting in each image. This is because, in the real-life world, some classes have a strong correlation, for example, cars are often driven or parked on pavements. To further demonstrate the class dependency, we calculate conditional probabilities for each of the two categories. Let $C_r$ denote referenced class, and $C_p$ denote potential co-occurrence class. Conditional probability $P(C_p|C_r)$, which depicts the possibility that $C_p$ exhibits in an image, where the existence of $C_r$ is priorly known, can be solved with Eq. \ref{eq:cp},

\begin{equation}
\label{eq:cp}
P(C_p|C_r) = \frac{P(C_p,C_r)}{P(C_r)}.
\end{equation}

$P(C_p,C_r)$ indicates the joint occurrence probability of $C_p$ and $C_r$, and $P(C_r)$ refers to the priori probability of $C_r$. Conditional probabilities of all class pairs in UCM multi-label datasets are listed in Fig. \ref{fig:cm_ucm}, and it is intuitive that some classes have strong dependencies. For instance, it is highly possible that there are pavements in images, which contain airplanes, buildings, cars, or tanks. Moreover, it is notable that class dependencies are not symmetric due to their particular properties. For example, $P(water|ship)$ is twice as $P(ship|water)$ due to the reason that the occurrence of ships always infer to the co-occurrence of water, while not vice versa. Therefore, to thoroughly dig out the correlation among various classes, it is crucial to model class probabilistic dependencies bidirectionally in a classification method.

To this end, we boil the multi-label classification down into a structured output problem, instead of a simple regression issue, and employ a unified framework of a CNN and a bidirectional RNN to 1) extract semantic features from raw images and 2) model image-label relations as well as bidirectional class dependencies, respectively.

\subsection{Network Architecture}
\begin{figure*}[!t]
\centering
\includegraphics[width=1\textwidth]{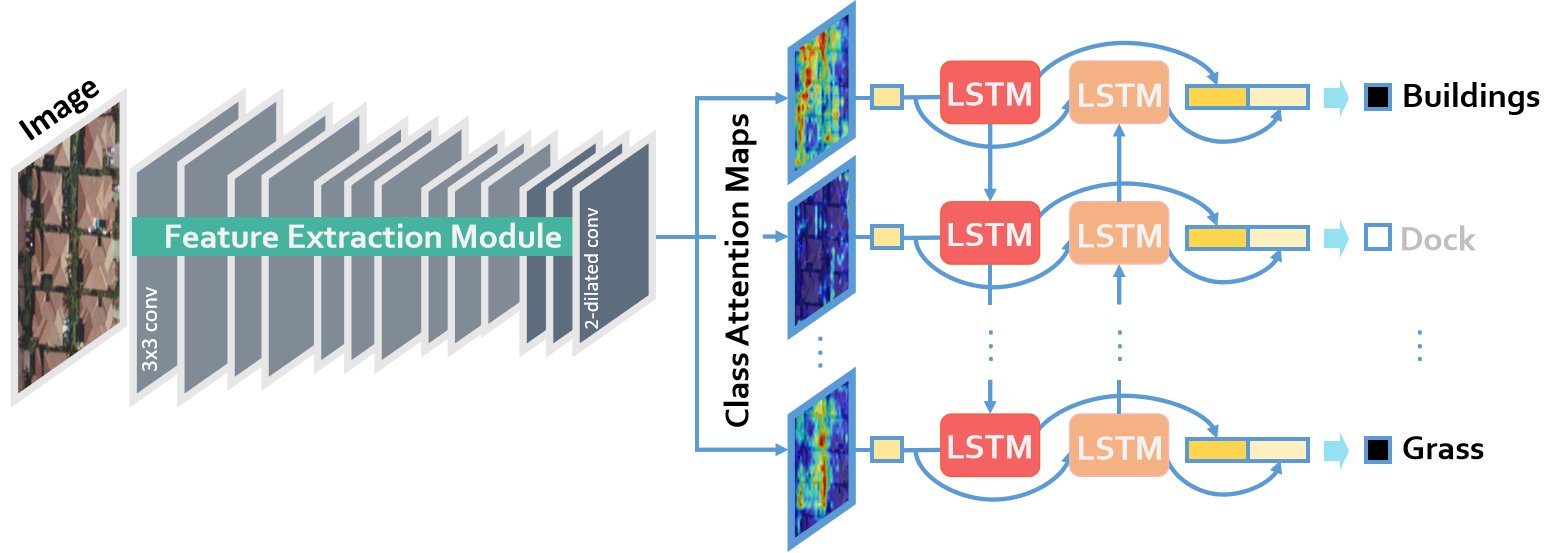}
\caption{The architecture of the proposed CA-Conv-BiLSTM for the multi-label classification of aerial images.}
\label{fig:network}
\end{figure*}

The proposed CA-Conv-BiLSTM, as illustrated in Fig. \ref{fig:network}, is composed of three components: a feature extraction module, a class attention learning layer, and a Bidirectional LSTM-based recurrent sub-network. More specifically, the feature extraction module employs a stack of interleaved convolutional and pooling layers to extract high-level features, which are then fed into a class attention learning layer to produce discriminative class-specific features. Afterwards, a bidirectional LSTM-based recurrent sub-network is attached to model both probabilistic class dependencies and underlying relationships between image features and labels.

Section \ref{sec:p1} details the architecture of the feature extraction module, and Section \ref{sec:p2} describes the explicit design of the class attention learning layer. Finally, Section \ref{sec:p3} introduces how to produce structured multi-label outputs from class-specific features via a bidirectional LSTM-based recurrent sub-network.

\subsubsection{Dense High-level Feature Extraction}
\label{sec:p1}

Learning efficient feature representations of input images is extremely crucial for the image classification task. To this end, a modern popular trend is to employ a CNN architecture to automatically extract discriminative features, and many recent studies \cite{hua2018lahnet, mou2018vehicle, xia2017aid, kang2018u, zhu2017deep, mou2017multitemporal} have achieved great progress in a wide range of classification tasks. Inspired by this, our model adapts VGG-16 \cite{simonyan2014very}, one of the most welcoming CNN architectures for its effectiveness and elegance, to extract high-level features for our task. 

Specifically, the feature extraction module consists of 5 convolutional blocks, and each of them contains 2  or 3 convolutional layers (as illustrated in the left of Fig. \ref{fig:network}). Notably, the number of filters is equivalent in a common convolutional block and doubles after each pooling layer, which is utilized to reduce the spatial dimension of feature maps. The purpose of such design is to enable the feature extraction module to learn diverse features at a less computational expense. The receptive field of all convolutional filters is $3 \times 3$, which increases nonlinearities inside the feature extraction module. Besides, the convolution stride is 1 pixel, and the spatial padding of each convolutional layer is set as 1 pixel as well. Among these convolutional blocks, max-pooling layers are interleaved for reducing the size of feature maps and meanwhile, maintaining only local representative, such as maximum in a $2 \times 2$-pixel region. The size of pooling windows is $2 \times 2$ pixels, and the pooling stride is 2 pixels, which halves feature maps in width and length. 

Features directly learned from a conventional CNN (e.g., VGG-16) are proved to be high-level and semantic, but their spatial resolution is significantly reduced, which is not favorable for generating high-dimensional class-specific features in the subsequent class attention learning layer. To address this, max-pooling layers following the last two convolutional blocks are discarded in our model, and atrous convolutional filters with dilation rate 2 are employed in the last convolutional block for preserving original receptive fields. Consequently, our feature extraction module is capable of learning high-level features with finer spatial resolution, so-called ``dense'', compared to VGG-16, and it is feasible to initialize our model with pre-trained VGG-16, considering that all filters have equivalent receptive fields. 

Moreover, it is noteworthy that other popular CNN architectures can be taken as prototypes of the feature extraction module, and thus, we extend researches to GoogLeNet \cite{szegedy2015going} and ResNet \cite{he2016deep} for a comprehensive evaluation of CA-Conv-BiLSTM. Regarding GoogLeNet, i.e., Inception-v3 \cite{szegedy2016rethinking}, the stride of convolutional and pooling layers after \textit{``mixed7''} is reduced to 1 pixel, and the dilation rate of convolutional filters in \textit{``mixed9''} is 2. For ResNet (we use ResNet-50), the convolution stride in last two residual blocks is set as 1 pixel, and the dilation rate of filters in the last residual block is 2. Besides, layers after global average pooling layers, as well as itself, are removed to ensure dense high-level feature maps.

\subsubsection{Class Attention Learning Layer}
\label{sec:p2}

Although Features extracted from pre-trained CNNs are high-level and can be directly fed into a fully connected layer for generating multi-label predictions, it is infeasible to learn high-order probabilistic dependencies by recurrently feeding it with identical features. Therefore, extracting discriminative class-wise features plays a key role in discovering class dependencies and effectively bridging CNN and RNN for multi-label classification tasks. 

\begin{figure*}[!t]
\centering
\subfloat[]{\includegraphics[width=1.1in]{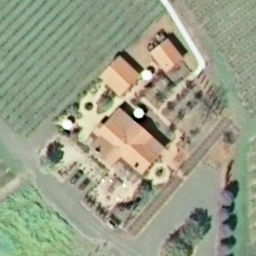}
\label{fig:ca1}}
\hspace{0.07in}
\subfloat[]{\includegraphics[width=1.1in]{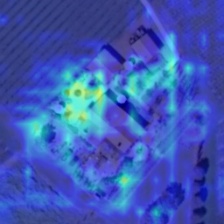}
\label{fig:ca2}}
\hspace{0.07in}
\subfloat[]{\includegraphics[width=1.1in]{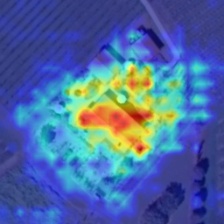}
\label{fig:ca3}}
\hspace{0.07in}
\subfloat[]{\includegraphics[width=1.1in]{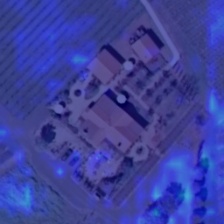}
\label{fig:ca0}}
\caption{Example class attention maps of an a) aerial image, with respect to different classes: b) bare soil, c) building, and d) water.}
\label{fig:ca}
\end{figure*}

Here, we propose a class attention learning layer to explore features with respect to each category, and the proposed layer, illustrated in the middle of Fig. \ref{fig:network}, consists of the following two stages: 1) generating class attention maps via a $1 \times 1$ convolutional layer with stride 1, and 2) vectorizing each class attention map to obtain class-specific features. Formally, given feature maps $\bm{X}$, extracted from the feature extraction module, with a size of $W \times W \times K$, and let $\bm{w}_l$ represent the $l$-th convolutional filter in the class attention learning layer. The attention map $\bm{M}_l$ for class $l$ can be obtained with the following formula:
\begin{equation}
\label{eq:ca}
\bm{M}_l = \bm{X} * \bm{w}_l,
\end{equation}
where $l$ ranges from 1 to the number of classes, and $*$ represents convolution operation. Considering that the size of convolutional filters is $1 \times 1$, a class attention map $\bm{M}_l$ is intrinsically a linear combination of all channels in $\bm{X}$. With this design, the proposed class attention learning layer is capable of learning discriminative class attention maps. Some examples are shown in Fig. \ref{fig:ca}. An aerial image (cf. Fig. \ref{fig:ca1}) in UCM multi-label dataset is first fed into the feature extraction module, adapted from VGG-16, and outputs of its last convolutional block are considered as the feature maps $\bm{X}$ in Eq. \ref{eq:ca}. Thus, $\bm{X}$ is abundant in high-level semantic information, and the size of $\bm{X}$ is $14 \times 14 \times 512$. Afterwards, a class attention learning layer, where the number of filters is equivalent to that of classes, is appended to generate class-specific feature representations with respect to all categories. With sufficient training, they are supposed to learn class-wise attention maps. It is observed that class attention maps highlight discriminative areas for different categories and exhibit almost no activations with respect to absent classes (as shown in Fig. \ref{fig:ca3}).  

Subsequently, class attention maps $\bm{M}_l$ are transformed into class-wise feature vectors $\bm{v}_l$ of $W^2$ dimensions by vectorization. Instead of fully connecting class attention maps to each hidden unit in the following layer, we construct class-wise connections between class attention maps and their corresponding hidden units, i.e., corresponding time steps in an LSTM layer in our network. In this way, features fed into different units are retained to be class-specific discriminative and significantly contribute to the exploitation of the dynamic class dependency in the subsequent bidirectional LSTM layer. 

\subsubsection{Class Dependency Learning via a BiLSTM-based Sub-network}
\label{sec:p3}

As an important branch of neural networks, RNN is widely used in dealing with sequential data, e.g., textual data and temporal series, due to its strong capabilities of exploiting implicit dependencies among inputs. Unlike CNN, RNN is characterized by its recurrent neurons, of which activations are dependent on both current inputs and previous hidden states. However, conventional RNNs suffer from the gradient vanishing problem and are found difficult to learn long-term dependencies. Therefore, in this work, we seek to model class dependencies with an LSTM-based RNN, which is first proposed in \cite{hochreiter1997long} and has shown great performance in processing long sequences \cite{graves2013generating, gers1999learning, xu2015show, Mournn, moulearn}.

Instead of directly summing up inputs as in a conventional recurrent layer, an LSTM layer relies on specifically designed hidden units, LSTM units, where information, such as the class dependency between category $l$ and $l-1$, is ``memorized'', updated, and transmitted with a memory cell and several gates. Specifically, given a class-specific feature $\bm{v}_l$ obtained from the class attention learning layer as an input of the LSTM memory cell $\bm{c}_l$ at time step $l$, and let $\bm{h}_l$ represent the activation of $\bm{c}_l$. New memory information $\tilde{\bm{c}}_l$, learned from the previous activation $\bm{h}_{l-1}$ and the present input feature $\bm{v}_l$, is obtained as follows:
\begin{equation}
\tilde{\bm{c}}_l = \tanh(\bm{W}_{cv} \bm{v}_l + \bm{W}_{ch}\bm{h}_{l-1}+\bm{b}_c),
\end{equation}
where $\bm{W}_{cv}$ and $\bm{W}_{ch}$ denote weight matrix from input vectors to memory cell and hidden-memory coefficient matrix, respectively, and $\bm{b}_c$ is a bias term. Besides, $\tanh(\cdot)$ is the hyperbolic tangent function. In contrast to conventional recurrent units, where the $\tilde{\bm{c}}_l$ is directly used to update the current state $\bm{h}_l$, an LSTM unit employs an input gate $\bm{i}_l$ to control the extent to which $\tilde{\bm{c}}_l$ is added, and meanwhile, partially omits uncorrelated prior information from $\bm{c}_{l-1}$ with a forget gate $\bm{f}_l$. The two gates are performed by the following equations:

\begin{equation}
\begin{array}{c}
\bm{i}_l = \sigma(\bm{W}_{iv}\bm{v}_l + \bm{W}_{ih} \bm{h}_{l-1} + \bm{W}_{ic}\bm{c}_{l-1} + \bm{b}_i), \vspace{0.5em}\\
\bm{f}_l = \sigma(\bm{W}_{fv}\bm{v}_l + \bm{W}_{fh} \bm{h}_{l-1} + \bm{W}_{fc}\bm{c}_{l-1} + \bm{b}_f).\vspace{0.5em}
\end{array}
\end{equation}

\begin{figure*}[!t]
\centering
\includegraphics[width=1\textwidth]{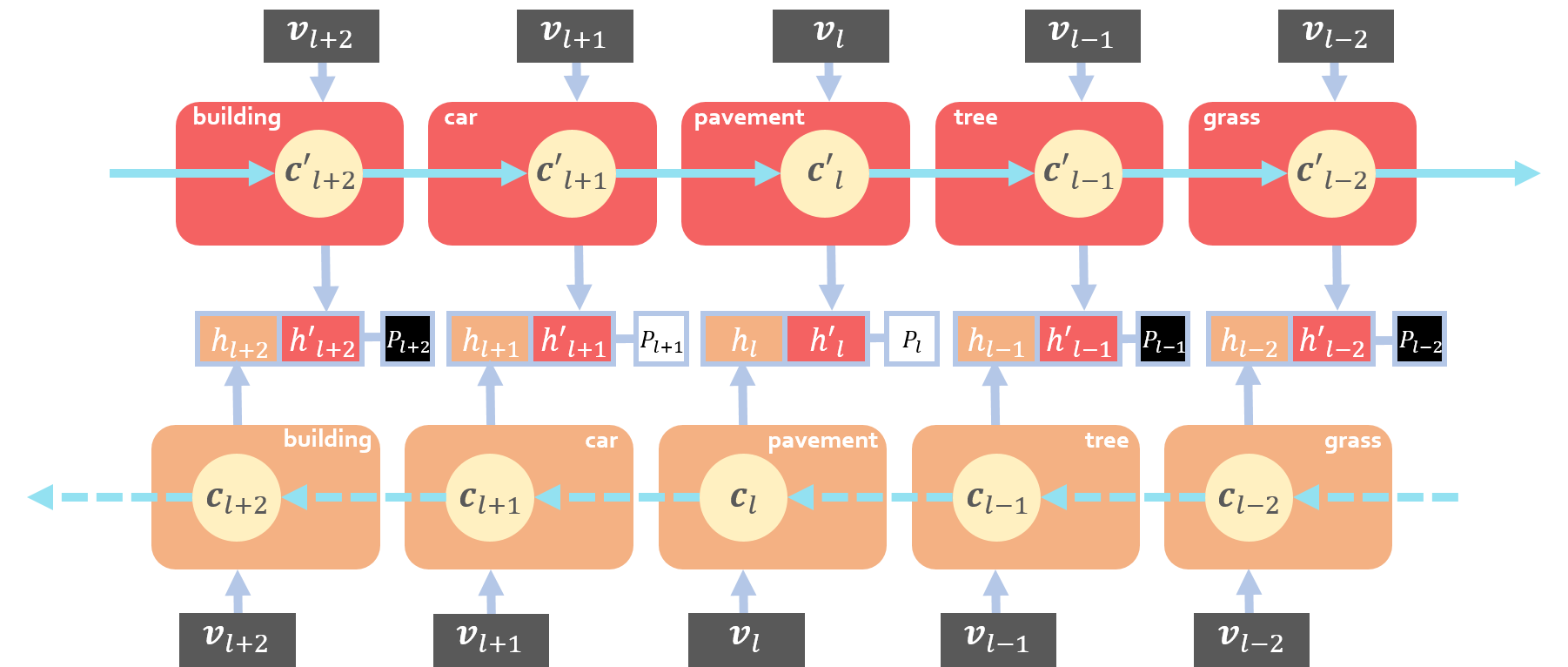}
\caption{Illustration of the bidirectional structure. The direction of the upper stream is opposite to that of the lower stream. Notably, $\bm{h'}_{l-1}$, $\bm{c'}_{l-1}$ denotes the activation and memory cell in the upper stream at the time step, which corresponds to class $l-1$ for convenience (considering that the subsequent time step is usually denoted as $l+1$).}
\label{fig:bilstm}
\end{figure*}

Consequently, the memory cell $\bm{c}_l$ is updated by
\begin{equation}
\bm{c}_l =  \bm{i}_l \odot \tilde{\bm{c}}_l + \bm{f}_l \odot \bm{c}_{l-1},
\end{equation}
where $\odot$ represents element-wise multiplication. Afterwards, an output gate $\bm{o}_l$, formulated by
\begin{equation}
\bm{o}_l =  \sigma(\bm{W}_{ov}\bm{v}_l + \bm{W}_{oh} \bm{h}_{l-1} + \bm{W}_{oc}\bm{c}_l+ \bm{b}_o),
\end{equation}
is designed to determine the proportion of memory content to be exposed, and eventually, the memory cell $\bm{c}_l$ at time step $l$ is activated by
\begin{equation}
\bm{h}_l = \bm{o}_l \tanh(\bm{c}_l).
\end{equation}

Although it is not difficult to discover that the activation of the memory cell at each time step is dependent on both input class-specific feature vectors and previous cell states. However, taking into account that the class dependency is bidirectional, as demonstrated in Section \ref{sec:ob}, a single-directional LSTM-based RNN is insufficient to draw a comprehensive picture of inter-class relevance. Therefore, a bidirectional LSTM-based RNN, composed of two identical recurrent streams but with reversed directions, is introduced in our model, and the hidden units are updated based on signals from not only their preceding states but also subsequent ones. 

In order to practically adapt a bidirectional LSTM-based RNN to modeling the class dependency, we set the number of time steps in our bidirectional LSTM-based sub-network equivalent to that of classes under the assumption that distinct classes are predicted at respective time steps. Validated in Section \ref{sec:out_ucm} and \ref{sec:out_dfc}, such design enjoys two outstanding characteristics: on one hand, the LSTM memory cell at time step $l$, $\bm{c}_l$, focuses on learning dependent relationship between class $l$ and others in dual directions (cf. Fig. \ref{fig:bilstm}), and on the other hand, the occurrence probability of class $l$, $P_l$, can be predicted from outputs $[\bm{h}_l, \bm{h}_l']$ with a single-unit fully connected layer:
\begin{equation}
P_l =  \sigma(\bm{w}_l[\bm{h}_l, \bm{h}_l']+\bm{b}_l),
\end{equation}
where $\bm{h}_l'$ denotes the activation of $\bm{c}_l$ in the other direction, and $\sigma$ is used as the activation function.

\section{Experiments and Discussion}
\label{sec:experiment}

In this section, two high-resolution aerial datasets of different resolution used for evaluating our network are first described in Section \ref{sec:data}, and then, the training strategies are introduced in Section \ref{sec:train}. Afterwards, the performance of the proposed network on the two datasets is quantitatively and qualitatively evaluated in the following sections.

\subsection{Data description}
\label{sec:data}
\subsubsection{UCM Multi-label Dataset}

\begin{figure*}[!t]
\centering
\subfloat[]{\includegraphics[width=0.13\textwidth]{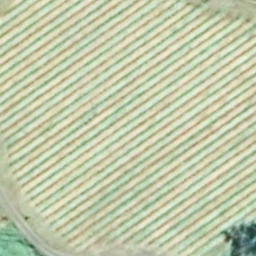}
\label{fig:s1}}
\hfil
\subfloat[]{\includegraphics[width=0.13\textwidth]{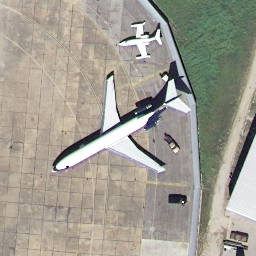}
\label{fig:s2}}
\hfil
\subfloat[]{\includegraphics[width=0.13\textwidth]{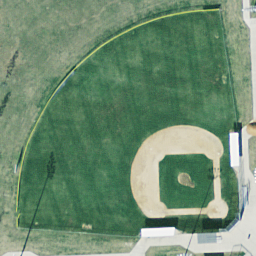}
\label{fig:s3}}
\hfil
\subfloat[]{\includegraphics[width=0.13\textwidth]{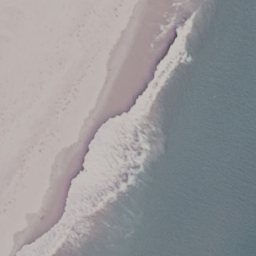}
\label{fig:s4}}
\hfil
\subfloat[]{\includegraphics[width=0.13\textwidth]{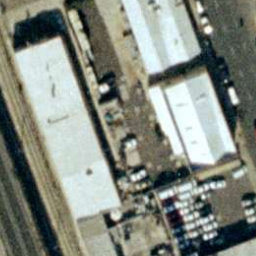}
\label{fig:s5}}
\hfil
\subfloat[]{\includegraphics[width=0.13\textwidth]{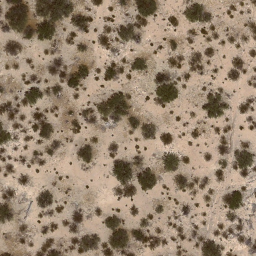}
\label{fig:s6}}
\hfil
\subfloat[]{\includegraphics[width=0.13\textwidth]{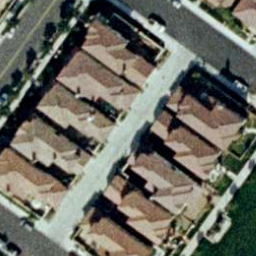}
\label{fig:s7}}
\vfill
\subfloat[]{\includegraphics[width=0.13\textwidth]{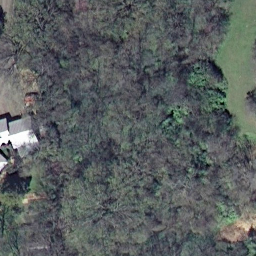}
\label{fig:s8}}
\hfil
\subfloat[]{\includegraphics[width=0.13\textwidth]{samples/freeway08}
\label{fig:s9}}
\hfil
\subfloat[]{\includegraphics[width=0.13\textwidth]{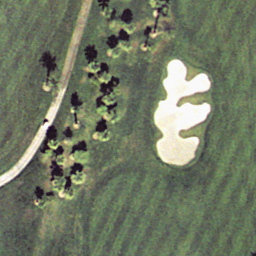}
\label{fig:s10}}
\hfil
\subfloat[]{\includegraphics[width=0.13\textwidth]{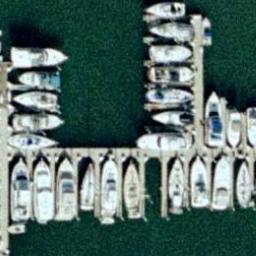}
\label{fig:s11}}
\hfil
\subfloat[]{\includegraphics[width=0.13\textwidth]{samples/intersection12}
\label{fig:s12}}
\hfil
\subfloat[]{\includegraphics[width=0.13\textwidth]{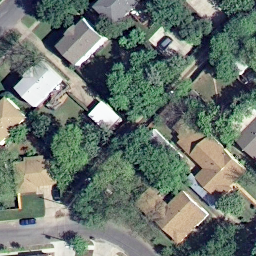}
\label{fig:s13}}
\hfil
\subfloat[]{\includegraphics[width=0.13\textwidth]{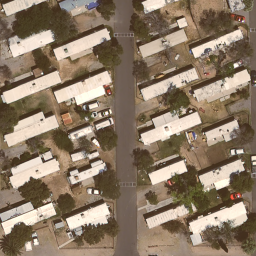}
\label{fig:s14}}
\vfill
\subfloat[]{\includegraphics[width=0.13\textwidth]{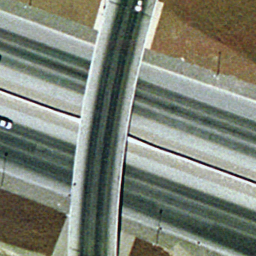}
\label{fig:s15}}
\hfil
\subfloat[]{\includegraphics[width=0.13\textwidth]{samples/parkinglot26}
\label{fig:s16}}
\hfil
\subfloat[]{\includegraphics[width=0.13\textwidth]{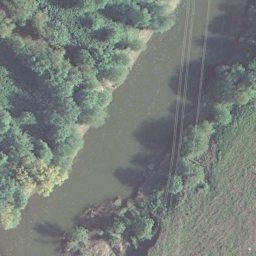}
\label{fig:s17}}
\hfil
\subfloat[]{\includegraphics[width=0.13\textwidth]{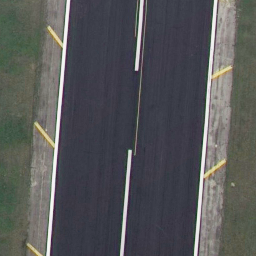}
\label{fig:s18}}
\hfil
\subfloat[]{\includegraphics[width=0.13\textwidth]{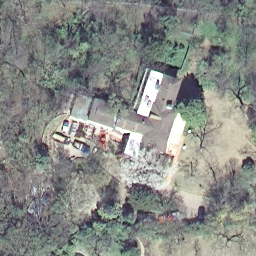}
\label{fig:s19}}
\hfil
\subfloat[]{\includegraphics[width=0.13\textwidth]{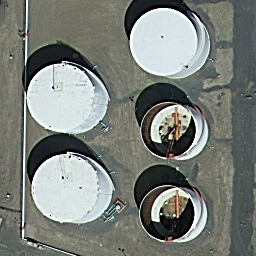}
\label{fig:s20}}
\hfil
\subfloat[]{\includegraphics[width=0.13\textwidth]{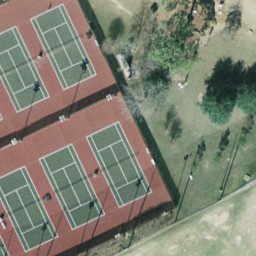}
\label{fig:s21}}
\caption{Example images from each scene category and their corresponding multiple \textit{object} labels in UCM multi-label dataset. Each image is $256 \times 256$ pixels with a spatial resolution of one foot, and their scene and \textit{object} labels are introduced: (a) Agricultural: \textsl{field} and \textsl{tree}. (b) Airplane: \textsl{airplane}, \textsl{bare soil}, \textsl{car}, \textsl{grass} and \textsl{pavement}. (c) Baseball diamond: \textsl{bare soil}, \textsl{building}, \textsl{grass}, and \textsl{pavement}. (d) Beach: \textsl{sand} and \textsl{sea}. (e) building: \textsl{building}, \textsl{car}, and \textsl{pavement}. (f) Chaparral: \textsl{bare soil} and \textsl{chaparral}. (g) Dense residential: \textsl{building}, \textsl{car}, \textsl{grass}, \textsl{pavement}, and \textsl{tree}. (h) Forest: \textsl{building}, \textsl{grass}, and \textsl{tree}. (i) Free way: \textsl{bare soil}, \textsl{car}, \textsl{grass}, \textsl{pavement}, and \textsl{tree}. (j) Golf course: \textsl{grass}, \textsl{pavement}, \textsl{sand}, and \textsl{tree}. (k) Harbor: \textsl{dock}, \textsl{ship}, and \textsl{water}. (l) Intersection: \textsl{building}, \textsl{car}, \textsl{grass}, \textsl{pavement}, and \textsl{tree}. (m) Medium residential: \textsl{building}, \textsl{car}, \textsl{grass}, \textsl{pavement}, and \textsl{tree}. (n) Mobile home park: \textsl{bare soil}, \textsl{car}, \textsl{grass}, \textsl{mobile home}, \textsl{pavement}, and \textsl{tree}. (o) Overpass: \textsl{bare soil}, \textsl{car}, and \textsl{pavement}. (p) Parking lot: \textsl{car}, \textsl{grass}, and \textsl{pavement}. (q) River: \textsl{grass}, \textsl{tree}, and \textsl{water}. (r) Runway: \textsl{grass} and \textsl{pavement}. (s) Sparse residential: \textsl{bare soil}, \textsl{building}, \textsl{car}, \textsl{grass}, \textsl{pavement}, and \textsl{tree}. (t) Storage tank: \textsl{bare soil}, \textsl{pavement}, and \textsl{tank}. (u) Tennis court: \textsl{bare soil}, \textsl{court}, \textsl{grass}, and \textsl{tree}.}
\label{fig:samp1}
\end{figure*}

UCM multi-label dataset \cite{chaudhuri2018multilabel} is reproduced from UCM dataset \cite{yang2010bag} by reassigning them with multiple object labels. Specifically, UCM dataset consists of 2100 aerial images of $256 \times 256$ pixels, and each of them is categorized into one of 21 scene labels: airplane, beach, agricultural, baseball diamond, building,  tennis courts, dense residential, forest, freeway, golf course, mobile home park, harbor, intersection, storage tank, medium residential, overpass, sparse residential, parking lot, river, runway, and chaparral. For each of them, there are 100 images with a spatial resolution of one foot collected by cropping manually from aerial ortho imagery provided by the United States Geological Survey (USGS) National Map.

In contrast, images in UCM multi-label dataset are relabeled by assigning each image sample with one or more labels based on their primitive objects. The total number of newly defined object classes is 17: airplane, sand, pavement, building, car, chaparral, court, tree, dock, tank, water, grass, mobile home, ship, bare soil, sea, and field. It is notable that several labels, namely, airplane, building, and tank, are defined in both datasets but with variant level. In UCM dataset, they are scene-level labels, since they are predominant objects in an image and used to depict the whole image, while in UCM multi-label dataset, they are object-level labels, regarded as candidate objects in a scene. The numbers of images related to each object category are listed in Table \ref{tab:u}, and examples from each scene category are shown in Fig. \ref{fig:samp1}, as well as their corresponding object labels. To train and test our network on UCM multi-label dataset, we select 80\% of sample images evenly from each scene category for training and the rest as the test set.

\begin{table}[!t]
\renewcommand{\arraystretch}{1.1}
\caption{The Number of Images in Each Object Class}
\label{tab:u}
\centering
\begin{tabular}{c|cccc}
\Xhline{3\arrayrulewidth}
Class No. & Class Name & Total & Training & Test \\
\hline
\hline
1 & airplane & 100 & 80 & 20 \\
2 & bare soil & 718 & 577 & 141 \\
3 & building & 691 & 555 & 136 \\
4 & car & 886 & 722 & 164 \\
5 & chaparral & 115 & 82 & 33 \\
6 & court & 105 & 84 & 21 \\
7 & dock & 100 & 80 & 20 \\
8 & field & 104 & 79 & 25\\
9 & grass & 975 & 804 & 171 \\
10 & mobile home & 102 & 82 & 20\\
11 & pavement & 1300 & 1047 & 253\\
12 & sand & 294 & 218 & 76 \\
13 & sea & 100 & 80 & 20 \\
14 & ship & 102 & 80 & 22 \\
15 & tank & 100 & 80 & 20 \\
16 & tree & 1009 & 801 & 208 \\
17 & water & 203 & 161 & 42\\
\hline
\hline
- & All & 2100 & 1680 & 420 \\
\Xhline{3\arrayrulewidth}
\end{tabular}
\end{table}

\subsubsection{DFC15 Multi-label Dataset}

Considering that images collected from the same scene may share similar patterns, alleviating task challenges, we build a new multi-label dataset, DFC15 multi-label dataset, based on a semantic segmentation dataset, DFC15 \cite{dfc15}, which was published and first used in 2015 IEEE GRSS Data Fusion Contest. DFC15 dataset is acquired over Zeebrugge with an airborne sensor, which is 300m off the ground. In total, 7 tiles are collected in DFC dataset, and each of them is $10000 \times 10000$ pixels with a spatial resolution of 5 cm. Unlike UCM dataset, where images are assigned with image-level labels, all tiles in DFC15 dataset are labeled in pixel-level, and each pixel is categorized into 8 distinct object classes: impervious, water, clutter, vegetation, building, tree, boat, and car. Notably, vegetation refers to low vegetation, such as bushes and grasses, and has no overlap with trees. Impervious indicates impervious surfaces (e.g., roads) excluding building rooftops.

\begin{figure*}[!t]
\centering
\subfloat[]{\includegraphics[width=0.23\textwidth]{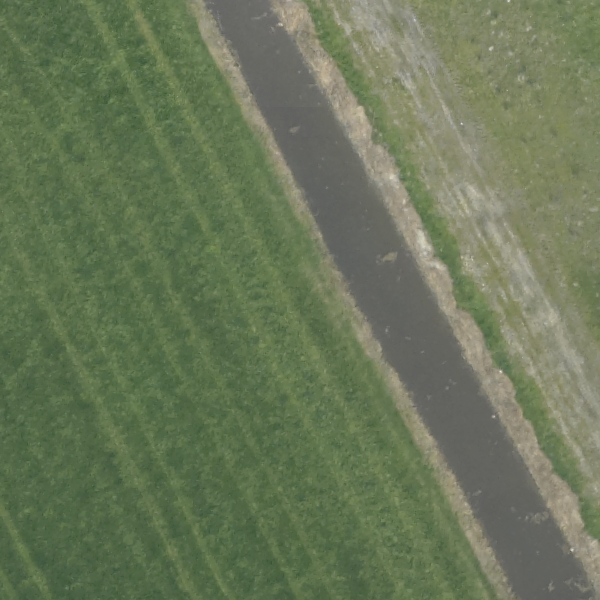}
\label{fig:d1}}
\hfil
\subfloat[]{\includegraphics[width=0.23\textwidth]{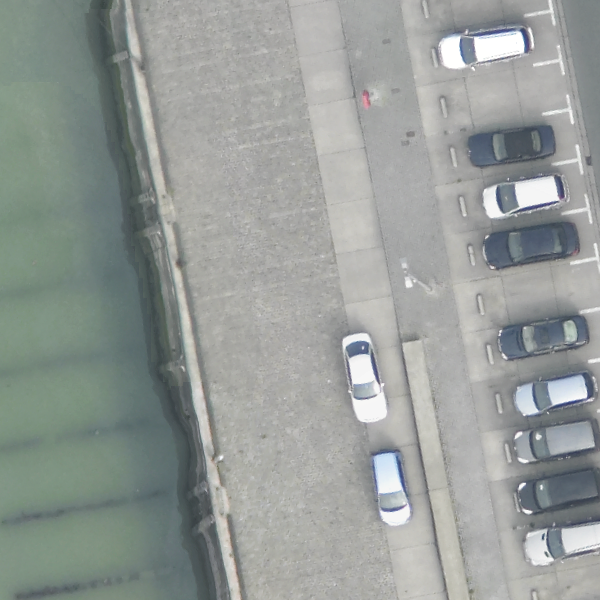}
\label{fig:d2}}
\hfil
\subfloat[]{\includegraphics[width=0.23\textwidth]{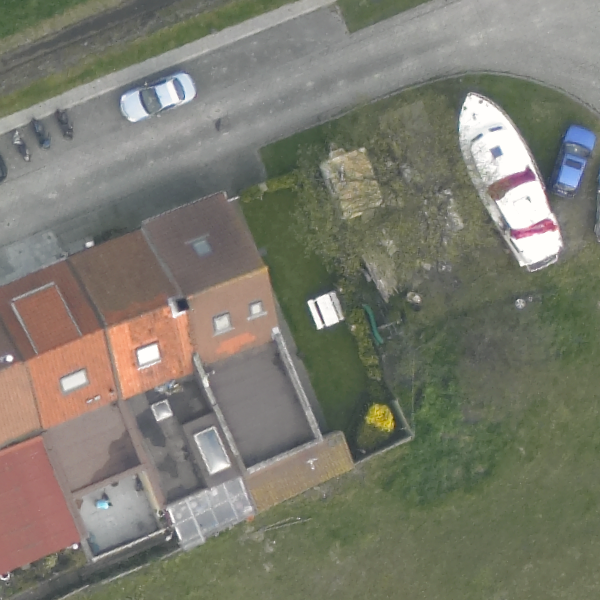}
\label{fig:d3}}
\hfil
\subfloat[]{\includegraphics[width=0.23\textwidth]{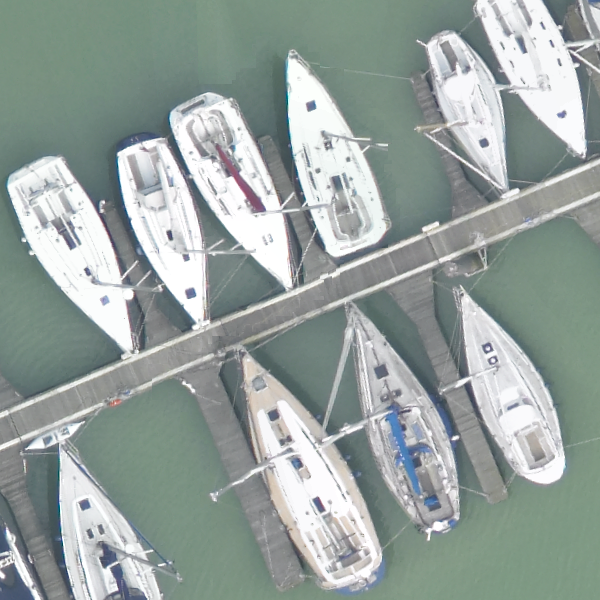}
\label{fig:d4}}
\vfil
\subfloat[]{\includegraphics[width=0.23\textwidth]{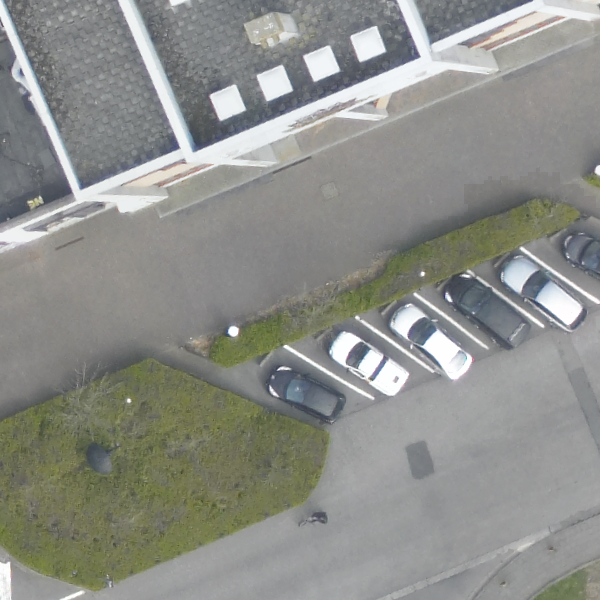}
\label{fig:d5}}
\hfil
\subfloat[]{\includegraphics[width=0.23\textwidth]{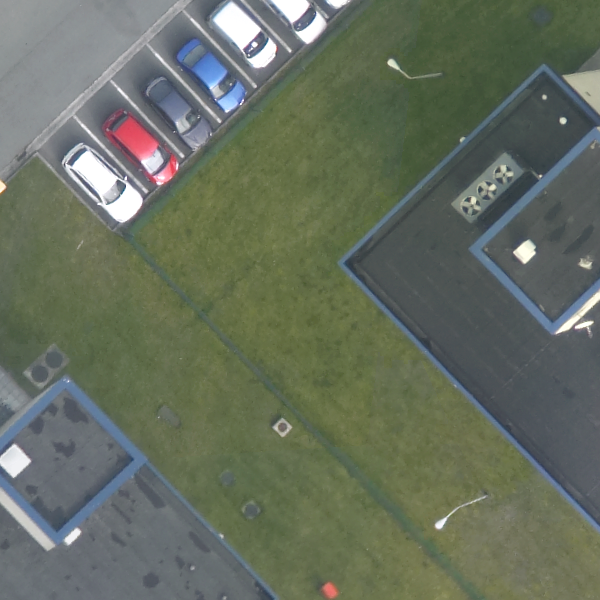}
\label{fig:d6}}
\hfil
\subfloat[]{\includegraphics[width=0.23\textwidth]{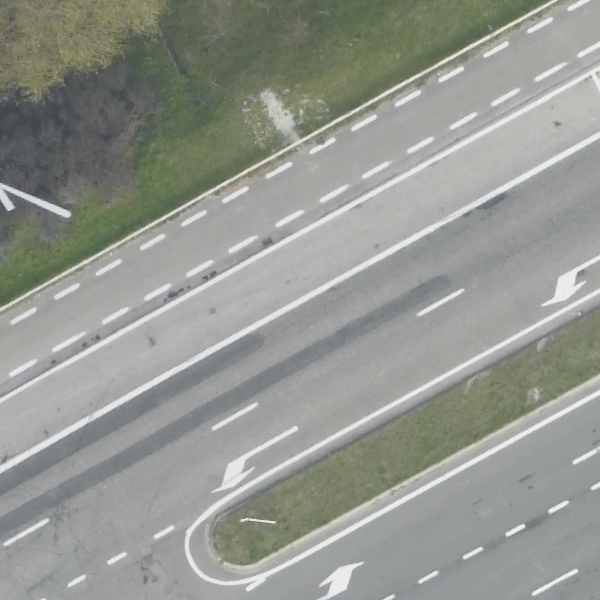}
\label{fig:d7}}
\hfil
\subfloat[]{\includegraphics[width=0.23\textwidth]{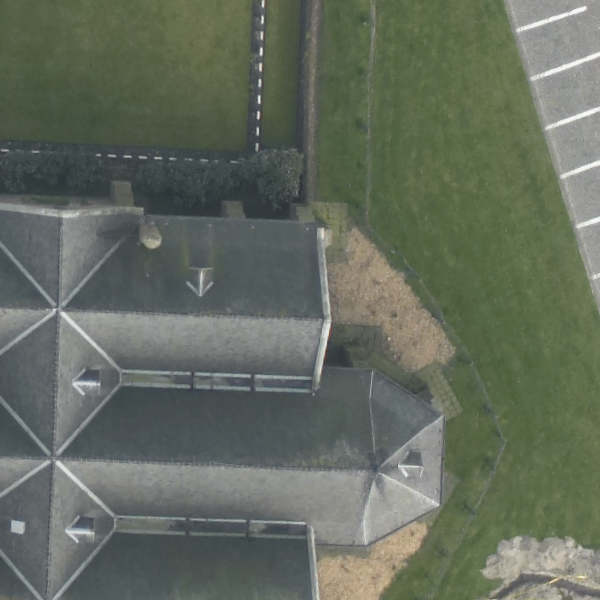}
\label{fig:d8}}
\caption{Example images in DFC15 multi-label dataset and their multiple \textit{object} labels. Each image is $600 \times 600$ pixels with a spatial resolution of 5 cm. (a) \textsl{Water} and \textsl{vegetation}. (b) \textsl{Impervious}, \textsl{water}, and \textsl{car}. (c) \textsl{Impervious}, \textsl{water}, \textsl{vegetation}, \textsl{building}, and \textsl{car}. (d) \textsl{Water}, \textsl{clutter}, and \textsl{boat}. (e) \textsl{Impervious}, \textsl{vegetation}, \textsl{building}, and \textsl{car}. (f) \textsl{Impervious}, \textsl{vegetation}, \textsl{building}, and \textsl{car}. (g) \textsl{Impervious}, \textsl{vegetation}, and \textsl{tree}. (h) \textsl{Impervious}, \textsl{vegetation}, and \textsl{building}.}
\label{fig:samp2}
\end{figure*}

Considering our task, the following processes are conducted: First, we crop large tiles into images of $600 \times 600$ pixels with a 200-pixel-stride sliding window. Afterwards, images containing unclassified pixels are ignored, and labels of all pixels in each image are aggregated into image-level multi-labels. An important characteristic of images in DFC15 multi-label dataset is lower inter-image similarity due to that they are cropped from vast regions consecutively without specific preferences, e.g., seeking images belonging to a specific scene. Moreover, extremely high resolution makes it more challenging as compared to UCM multi-label dataset. The numbers of images containing each object label are listed in Table \ref{tab:d}, and example images with their image-level object labels are shown in Fig. \ref{fig:samp2}. To conduct the evaluation, 80\% of images are randomly selected as the training set, while the others are utilized to test our network.

\begin{table}[!t]
\renewcommand{\arraystretch}{1.1}
\caption{The Number of Images in Each Object Class}
\label{tab:d}
\centering
\begin{tabular}{c|cccc}
\Xhline{3\arrayrulewidth}
Class No. & Class Name & Total & Training & Test \\
\hline
\hline
1 & impervious & 3133 & 2532 & 602 \\
2 & water & 998 & 759 & 239 \\
3 & clutter & 1891 & 1801 & 90 \\
4 & vegetation & 1086 & 522 & 562 \\
5 & building & 1001 & 672 & 330 \\
6 & tree & 258 & 35 & 223 \\
7 & boat & 270 & 239 & 31 \\
8 & car & 705 & 478 & 277 \\
\hline
\hline
- & All & 3342 & 2674 & 668 \\
\Xhline{3\arrayrulewidth}
\end{tabular}
\end{table}

\subsection{Training details}
\label{sec:train}
The proposed CA-Conv-BiLSTM is initialized with separate strategies with respect to three dominant components: 1) the feature extraction module is initialized with CNNs pre-trained on ImageNet dataset \cite{imagenet_cvpr09}, 2) convolutional filters in the class attention learning layer is initialized with a Glorot uniform initializer, and 3) all weights in the bidirectional 2048-d LSTM layer are randomly initialized in the range of $[-0.1,0.1]$ with a uniform distribution. Notably, weights in the feature extraction module are trainable and fine-tuned during the training phase of our network.

Regarding the optimizer, we chose Adam with Nesterov momentum \cite{nadam2}, claimed to converge faster than stochastic gradient descent (SGD), and set parameters of the optimizer as recommended: $\beta_1 = 0.9$, $\beta_2 = 0.999$, and $\epsilon = 1e-08$. The learning rate is set as $1e-04$ and decayed by 0.1 when the validation accuracy is saturated. The loss of the network is defined as the binary cross entropy. We implement the network on TensorFlow and train it on one NVIDIA Tesla P100 16GB GPU for 100 epochs. The size of the training batch is 32 as a trade-off between GPU memory capacity and training speed. To avoid overfitting, we stop training procedure when the loss fails to decrease in five epochs. Concerning ground truths, multiple labels of an image are encoded into a multi-hot binary sequence, of which the length is equivalent to the number of all candidate labels. For each digit, 1 indicates the existence of its corresponding label, while 0 denotes the absent label.

\subsection{Results on UCM Multi-label Dataset}
\label{sec:out_ucm}
\subsubsection{Quantitative Results}
To evaluate the performance of CA-Conv-BiLSTM for multi-label classification of high resolution aerial imagery, we calculate both $F_1$ \cite{wu2016unified} and $F_2$ \cite{kaggleplanet} score as follows:
\begin{equation}
\label{eq:f2}
F_\beta = (1+\beta^2)\frac{p_er_e}{\beta^2p_e+r_e}, \hspace{1em} \beta=1, 2,
\end{equation}
where $p_e$ is the example-based precision \cite{tsoumakas2007random} of predicted multiple labels, and $r_e$ indicates the example-based recall. They are computed by:
\begin{equation}
p_e = \frac{TP_e}{TP_e+FP_e},\hspace{1em} r_e = \frac{TP_e}{TP_e+FN_e},
\end{equation}
where $TP_e$, $FP_e$, and $FN_e$ indicate the numbers of positive labels, which are predicted correctly (true positives) and incorrectly (false positives), and negative labels, which are incorrectly predicted (false negatives) in an example (i.e., an image with multiple object labels in our case), respectively. Then, the average of $F_2$ scores of each example is formed to assess the overall accuracy of multi-label classification tasks. Besides, example-based mean precision as well as mean recall are calculated to assess the performance from the perspective of examples, while label-based mean precision and mean recall can help us understand the performance of the network from the perspective of object labels:
\begin{equation}
p_l = \frac{TP_l}{TP_l+FP_l},\hspace{1em} r_l = \frac{TP_l}{TP_l+FN_l},
\end{equation}
where $TP_l$, $FP_l$, and $FN_l$ represent the numbers of correctly predicted positive images, incorrectly predicted positive images, and incorrectly predicted negative images with respect to each label. 

For a fair validation of CA-Conv-BiLSTM, we decompose the evaluation into two components: we compare 1) CA-Conv-LSTM with standard CNNs to validate the effectiveness of employing LSTM-based recurrent sub-network, and 2) CA-Conv-BiLSTM with CA-Conv-LSTM for further assess the significance of the bidirectional structure. The detailed configurations of these competitors are listed in Table \ref{tab:nets}. For standard CNNs, we substitute last softmax layers, which are designed for single-label classification, with sigmoid layers to predict multi-hot binary sequences, where each digit indicates the probability of the presence of its corresponding category. To calculate evaluation metrics, we binarize outputs of all models with a threshold of 0.5 for producing binary sequences. Besides, our model is compared with a relevant existing method \cite{zeggada2017deep} for a comprehensive evaluation of its performance.
\begin{table}[t]
\centering
\renewcommand{\arraystretch}{1.3}
\begin{threeparttable}
\caption{Configurations of CA-Conv-LSTM Architectures}
\label{tab:nets}
\begin{tabular}{cccc}
\Xhline{3\arrayrulewidth}
Model & CNN model & Class Attention Map & Bi.\\
\hline
CA-VGG-LSTM & VGG-16 & $28 \times 28 \times N$ & \ding{55} \\
\textbf{CA-VGG-BiLSTM} & VGG-16 & $28 \times 28 \times N$ & \checkmark \\
\hline
\hline
CA-GoogLeNet-LSTM & Inception-v3 & $17 \times 17 \times N$ & \ding{55} \\
\textbf{CA-GoogLeNet-BiLSTM} & Inception-v3 & $17 \times 17 \times N$ & \checkmark \\
\hline
\hline
CA-ResNet-LSTM & ResNet-50 & $28 \times 28 \times N$ & \ding{55} \\
\textbf{CA-ResNet-BiLSTM} & ResNet-50 & $28 \times 28 \times N$ & \checkmark \\
\Xhline{3\arrayrulewidth}
\end{tabular}
\begin{tablenotes}
\item[] N indicates the number of classes in the dataset.
\item[] Bi. indicates whether the model is bidirectional or not.
\end{tablenotes}
\end{threeparttable}
\end{table}

Table \ref{tab:ucm} exhibits results on UCM multi-label dataset, and it can be seen that compared to directly applying standard CNNs to multi-label classification, CA-Conv-LSTM framework performs superiorly as expected due to taking class dependencies into consideration. CA-VGG-LSTM increases the mean $F_1$ score by 1.03\% with respect to VGGNet, while for CA-ResNet-LSTM, an increment of 1.68\%, is obtained compared to ResNet. Mostly enjoying this framework, CA-GoogLeNet-LSTM achieves the best mean $F_1$ score of 81.78\% and an increment of 1.10\% in comparison with other CA-Conv-LSTM models and GoogLeNet, respectively. Moreover, CA-ResNet-LSTM shows an improvement of 3.08\% of the mean $F_2$ score in comparison with ResNet, while CA-GoogLeNet-LSTM obtains the best $F_2$ score of 85.16\%. To summarize, all comparisons demonstrate that instead of directly using a standard CNN as a regression task, exploiting class dependencies plays a key role in multi-label classification.

\begin{table}[t]
\renewcommand{\arraystretch}{1.3}
\caption{Quantitative Results on UCM Multi-label Dataset (\%)}
\label{tab:ucm}
\centering
\begin{threeparttable}
\begin{tabular}{ccccccc}
\Xhline{3\arrayrulewidth}
Model & m.$F_1$ & m.$F_2$ & m.$\text{P}_{e}$ & m.$\text{R}_{e}$ & m.$\text{P}_{l}$ & m.$\text{R}_{l}$\\
\hline
VGGNet \cite{simonyan2014very} & 78.54 & 80.17 & 79.06 & 82.30 & 86.02 & 80.21 \\
VGG-RBFNN \cite{zeggada2017deep} & 78.80 & 81.14 & 78.18 & 83.91 & 81.90 & 82.63\\
CA-VGG-LSTM & 79.57 & 80.75 & 80.64 & 82.47 & 87.74 & 75.95 \\
\textbf{CA-VGG-BiLSTM} & \textbf{79.78} & \textbf{81.69} & 79.33 & 83.99 & 85.28 & 76.52 \\
\hline
\hline
GoogLeNet \cite{szegedy2015going} & 80.68 & 82.32 & 80.51 & 84.27 & 87.51 & 80.85 \\
GoogLeNet-RBFNN \cite{zeggada2017deep} & 81.54 & 84.05 & 79.95 & 86.75 & 86.19 & 84.92\\ 
CA-GoogLeNet-LSTM & 81.78 & \textbf{85.16} & 78.52 & 88.60 & 86.66 & 85.99 \\
\textbf{CA-GoogLeNet-BiLSTM} & \textbf{81.82} & 84.41 & 79.91 & 87.06 & 86.29 & 84.38\\
\hline
\hline
ResNet-50 \cite{he2016deep} & 79.68 & 80.58 & 80.86 & 81.95 & 88.78 & 78.98 \\
ResNet-RBFNN \cite{zeggada2017deep} & 80.58 & 82.47 & 79.92 & 84.59 & 86.21 & 83.72 \\
CA-ResNet-LSTM & 81.36 & 83.66 & 79.90 & 86.14 & 86.99 & 82.24\\
\textbf{CA-ResNet-BiLSTM} & \textbf{81.47} & \textbf{85.27} & 77.94 & 89.02 & 86.12 & 84.26\\
\Xhline{3\arrayrulewidth}
\end{tabular}

\begin{tablenotes}
\item[] m.$F_1$ and m.$F_2$ indicate the mean $F_1$ and $F_2$ score.
\item[] m.$\text{P}_e$ and m.$\text{R}_e$ indicate mean example-based precision and recall.
\item[] m.$\text{P}_l$ and m.$\text{R}_l$ indicate mean label-based precision and recall.
\end{tablenotes}
\end{threeparttable}
\end{table}

\begin{table}
\caption{Example Predictions on UCM and DFC15 Multi-label Dataset}
\label{tab:output}
\centering
\begin{adjustwidth}{-2.95cm}{0cm}
\renewcommand{\arraystretch}{1.3}
\begin{threeparttable}
\begin{tabular}{>{\centering\arraybackslash}m{2.9cm}>{\centering\arraybackslash}m{2.9cm}>{\centering\arraybackslash}m{2.9cm}>{\centering\arraybackslash}m{2.9cm}>{\centering\arraybackslash}m{2.9cm}>{\centering\arraybackslash}m{2.9cm}} 
\Xhline{3\arrayrulewidth}
Images in UCM Multi-label Dataset 
& \subfloat{\includegraphics[width=0.19\textwidth]{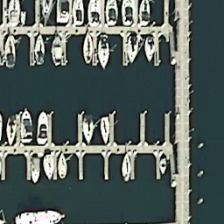}}
& \subfloat{\includegraphics[width=0.19\textwidth]{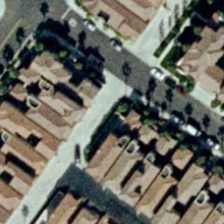}}
& \subfloat{\includegraphics[width=0.19\textwidth]{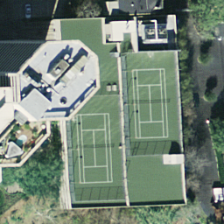}}
& \subfloat{\includegraphics[width=0.19\textwidth]{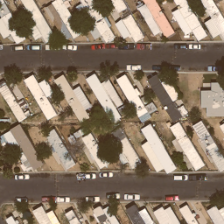}}
& \subfloat{\includegraphics[width=0.19\textwidth]{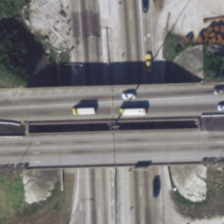}} \\
\hline
Ground Truths & dock, ship, and water & building, car, pavement, and tree & building, court, pavement, grass, and tree & car, pavement, mobile-home, and tree & bare soil, car, grass, and pavement \\
\hline
Predictions & dock, ship, and water & building, car, pavement, and tree & building, court, pavement, grass, and tree & car, pavement, mobile-home, tree and  \textcolor{red}{grass} & \textcolor{blue}{bare soil}, car, grass, \textcolor{red}{tree}, and pavement \\
\hline 
\hline
Images in DFC15 Multi-label Dataset 
& \subfloat{\includegraphics[width=0.19\textwidth]{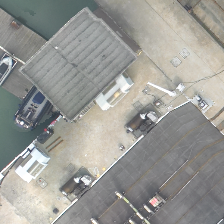}}
& \subfloat{\includegraphics[width=0.19\textwidth]{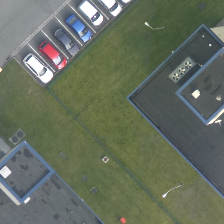}}
& \subfloat{\includegraphics[width=0.19\textwidth]{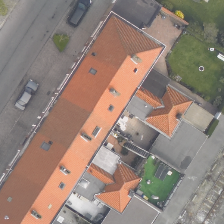}}
& \subfloat{\includegraphics[width=0.19\textwidth]{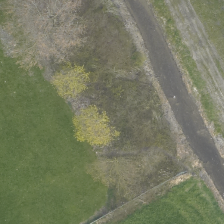}}
& \subfloat{\includegraphics[width=0.19\textwidth]{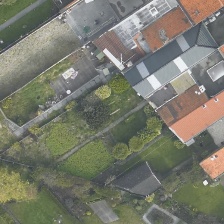}} \\
\hline
Ground Truths & impervious, water, and building & impervious, vegetation, car, and building & impervious, vegetation, building, clutter, and car & water, vegetation, tree & impervious, vegetation, building, car, and tree \\
\hline
Predictions & impervious, water, and building & impervious, vegetation, car, and building & impervious, vegetation, building, clutter, and car & \textcolor{red}{impervious}, \textcolor{blue}{water}, \textcolor{blue}{tree}, vegetation & impervious, vegetation, building, car, and \textcolor{blue}{tree} \\
\Xhline{3\arrayrulewidth}
\end{tabular}
\begin{tablenotes}
\item[] \textcolor{red}{Red} predictions indicate false positives, while \textcolor{blue}{blue} predictions are false negatives.
\end{tablenotes}
\end{threeparttable}
\end{adjustwidth}
\end{table}

Concerning the signification of employing a bidirectional structure, CA-Conv-BiLSTM performs better than CA-Conv-LSTM in the mean $F_1$ score, and compared to Conv-RBFNN, our models achieve higher mean $F_1$ and $F_2$ scores, increased by at most 0.98\% and 2.80\%, respectively. Another important observation is that our proposed model is equipped with higher example-based recall but lower example-based precision, which leads to a relatively higher mean $F_2$ score. Notably, the $F_2$ score is an evaluation index used in Kaggle Amazon contest \cite{kaggleplanet} to assess the performance of recognizing challenging rare objects in aerial images, and a higher score indicates a stronger capability. Table \ref{tab:output} exhibits several example predictions in UCM multi-label dataset. Although our model successfully predicts most multiple object labels, it is observed that the grass and tree are prone to be misclassified due to their analogous appearances. In the 4th image, the grass is a false positive when there exist trees, while in the 5th image, the tree is a false positive when the grass presents. Likewise, the bare soil in the 5th image is neglected unfortunately for its similar visual patterns with the grass.

\begin{figure*}[!t]
\captionsetup[subfigure]{labelformat=empty}
\centering
\subfloat{\includegraphics[width=0.102\textwidth]{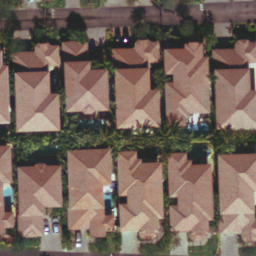}
\label{fig:v1}}
\hspace{-0.4em}
\subfloat{\includegraphics[width=0.102\textwidth]{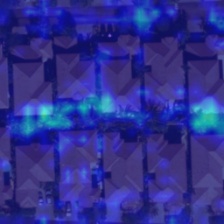}
\label{fig:v2}}
\hspace{-0.4em}
\subfloat{\includegraphics[width=0.102\textwidth]{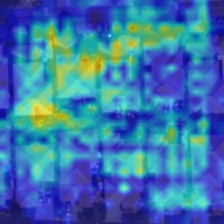}
\label{fig:v3}}
\hspace{-0.4em}
\subfloat{\includegraphics[width=0.102\textwidth]{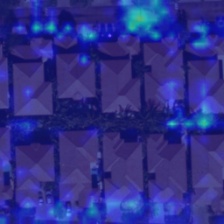}
\label{fig:v4}}
\hspace{-0.4em}
\subfloat{\includegraphics[width=0.102\textwidth]{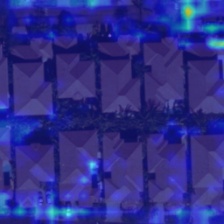}
\label{fig:v5}}
\hspace{-0.4em}
\subfloat{\includegraphics[width=0.102\textwidth]{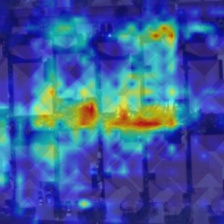}
\label{fig:v6}}
\hspace{-0.4em}
\subfloat{\includegraphics[width=0.102\textwidth]{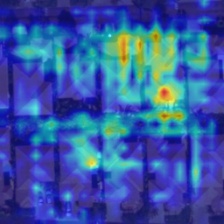}
\label{fig:v7}}
\hspace{-0.4em}
\subfloat{\includegraphics[width=0.102\textwidth]{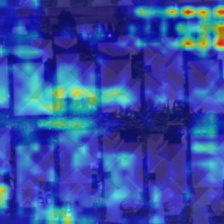}
\label{fig:v8}}
\hspace{-0.4em}
\subfloat{\includegraphics[width=0.102\textwidth]{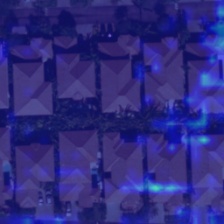}
\label{fig:v9}}
\vspace{-0.5em}
\subfloat{\includegraphics[width=0.102\textwidth]{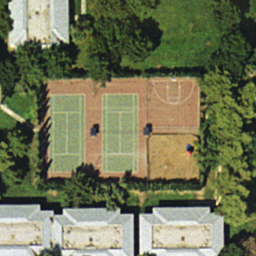}
\label{fig:v10}}
\hspace{-0.4em}
\subfloat{\includegraphics[width=0.102\textwidth]{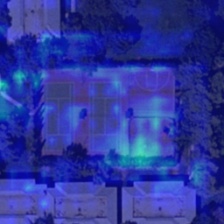}
\label{fig:v11}}
\hspace{-0.4em}
\subfloat{\includegraphics[width=0.102\textwidth]{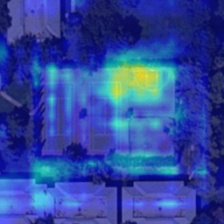}
\label{fig:v12}}
\hspace{-0.4em}
\subfloat{\includegraphics[width=0.102\textwidth]{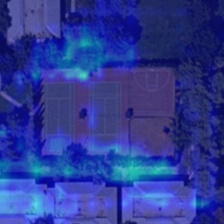}
\label{fig:v13}}
\hspace{-0.4em}
\subfloat{\includegraphics[width=0.102\textwidth]{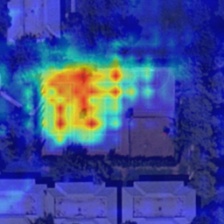}
\label{fig:v14}}
\hspace{-0.4em}
\subfloat{\includegraphics[width=0.102\textwidth]{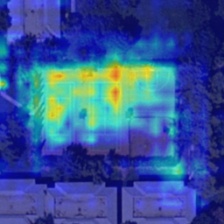}
\label{fig:v15}}
\hspace{-0.4em}
\subfloat{\includegraphics[width=0.102\textwidth]{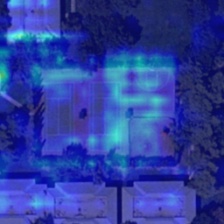}
\label{fig:v16}}
\hspace{-0.4em}
\subfloat{\includegraphics[width=0.102\textwidth]{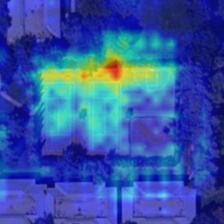}
\label{fig:v17}}
\hspace{-0.4em}
\subfloat{\includegraphics[width=0.102\textwidth]{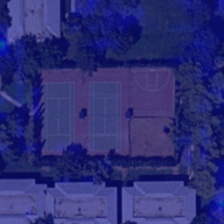}
\label{fig:v18}}
\vspace{-0.5em}
\subfloat{\includegraphics[width=0.102\textwidth]{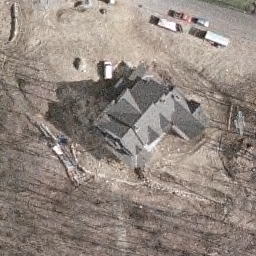}
\label{fig:v19}}
\hspace{-0.4em}
\subfloat{\includegraphics[width=0.102\textwidth]{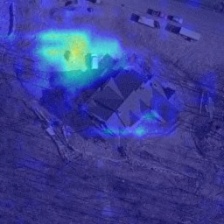}
\label{fig:v20}}
\hspace{-0.4em}
\subfloat{\includegraphics[width=0.102\textwidth]{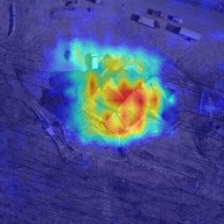}
\label{fig:v21}}
\hspace{-0.4em}
\subfloat{\includegraphics[width=0.102\textwidth]{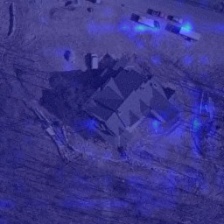}
\label{fig:v22}}
\hspace{-0.4em}
\subfloat{\includegraphics[width=0.102\textwidth]{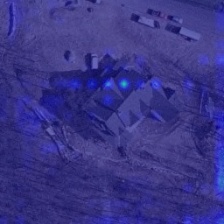}
\label{fig:v23}}
\hspace{-0.4em}
\subfloat{\includegraphics[width=0.102\textwidth]{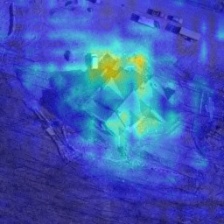}
\label{fig:v24}}
\hspace{-0.4em}
\subfloat{\includegraphics[width=0.102\textwidth]{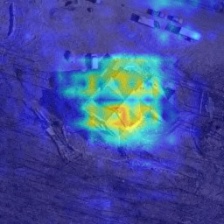}
\label{fig:v25}}
\hspace{-0.4em}
\subfloat{\includegraphics[width=0.102\textwidth]{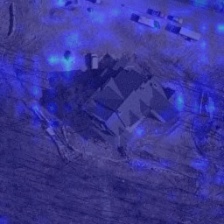}
\label{fig:v26}}
\hspace{-0.4em}
\subfloat{\includegraphics[width=0.102\textwidth]{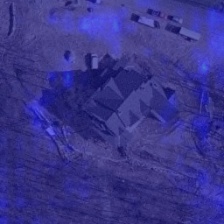}
\label{fig:v27}}
\vspace{-0.5em}
\subfloat{\includegraphics[width=0.102\textwidth]{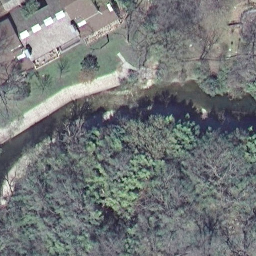}
\label{fig:v28}}
\hspace{-0.4em}
\subfloat{\includegraphics[width=0.102\textwidth]{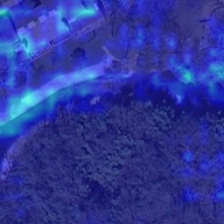}
\label{fig:v29}}
\hspace{-0.4em}
\subfloat{\includegraphics[width=0.102\textwidth]{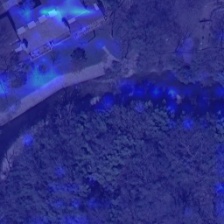}
\label{fig:v30}}
\hspace{-0.4em}
\subfloat{\includegraphics[width=0.102\textwidth]{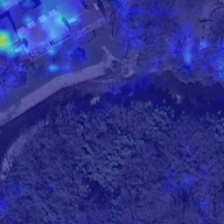}
\label{fig:v31}}
\hspace{-0.4em}
\subfloat{\includegraphics[width=0.102\textwidth]{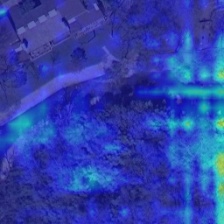}
\label{fig:v32}}
\hspace{-0.4em}
\subfloat{\includegraphics[width=0.102\textwidth]{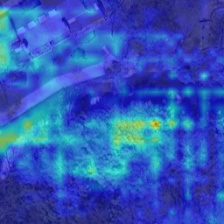}
\label{fig:v33}}
\hspace{-0.4em}
\subfloat{\includegraphics[width=0.102\textwidth]{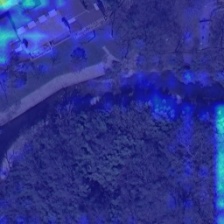}
\label{fig:v34}}
\hspace{-0.4em}
\subfloat{\includegraphics[width=0.102\textwidth]{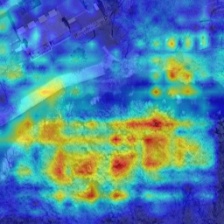}
\label{fig:v35}}
\hspace{-0.4em}
\subfloat{\includegraphics[width=0.102\textwidth]{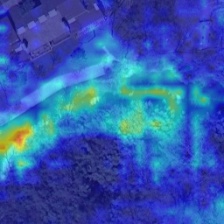}
\label{fig:v36}}
\vspace{-0.5em}
\subfloat[(a)]{\includegraphics[width=0.102\textwidth]{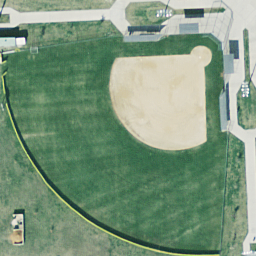}
\label{fig:v37}}
\hspace{-0.4em}
\subfloat[(b)]{\includegraphics[width=0.102\textwidth]{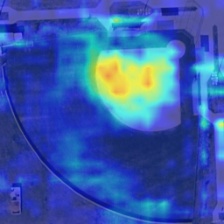}
\label{fig:v38}}
\hspace{-0.4em}
\subfloat[(c)]{\includegraphics[width=0.102\textwidth]{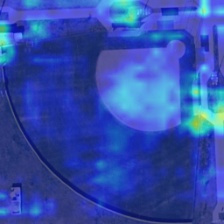}
\label{fig:v39}}
\hspace{-0.4em}
\subfloat[(d)]{\includegraphics[width=0.102\textwidth]{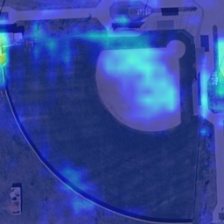}
\label{fig:v40}}
\hspace{-0.4em}
\subfloat[(e)]{\includegraphics[width=0.102\textwidth]{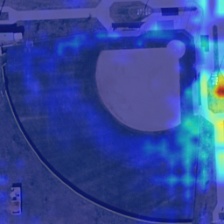}
\label{fig:v41}}
\hspace{-0.4em}
\subfloat[(f)]{\includegraphics[width=0.102\textwidth]{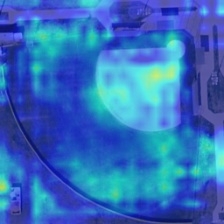}
\label{fig:v42}}
\hspace{-0.4em}
\subfloat[(g)]{\includegraphics[width=0.102\textwidth]{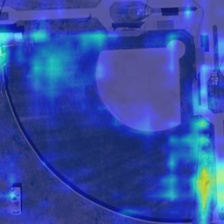}
\label{fig:v43}}
\hspace{-0.4em}
\subfloat[(h)]{\includegraphics[width=0.102\textwidth]{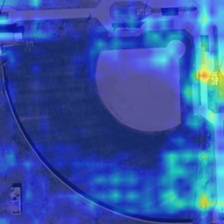}
\label{fig:v44}}
\hspace{-0.4em}
\subfloat[(i)]{\includegraphics[width=0.102\textwidth]{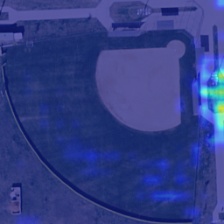}
\label{fig:v45}}
\caption{Example class attention maps of (a) images in UCM multi-label dataset with respect to (b) bare soil, (c) building, (d) car, (e) court, (f) grass, (g) pavement, (h) tree, and (i) water. Red indicates strong activations, while blue represents non-activations. Besides, normalization is performed based on each row for a fair comparison among class attention maps of the same images.}
\label{fig:visual1}
\end{figure*}

\subsubsection{Qualitative Results}

In addition to validate classification capabilities of the network by computing the mean $F_2$ score, we further explore the effectiveness of class-specific features learned from the proposed class attention learning layer and try to``open'' the black box of our network by feature visualization. Example class attention maps produced by the proposed network on UCM multi-label dataset are shown in Fig. \ref{fig:visual1}, where column (a) is original images, and columns (b)-(i) are class attention maps for different objects: (b) bare soil, (c) building, (d) car, (e) court, (f) grass, (g) pavement, (h) tree, and (i) water.  As we can see, these maps highlight discriminative regions for positive classes, while present almost no activations when corresponding objects are absent in original images. For example, object labels of the image at the first row in Fig. \ref{fig:visual1} are building, grass, pavement, and tree, and its class attention maps for these categories are strongly activated. From images at the fourth row of Fig. \ref{fig:visual1}, it can be seen that regions of the grassland, forest, and river are highlighted in their corresponding class attention maps, leading to positive predictions, while no discriminative areas are intensively activated in the other maps.

\subsection{Results on DFC15 Multi-label Dataset}
\label{sec:out_dfc}
\subsubsection{Quantitative Results}

\begin{table}[t]
\renewcommand{\arraystretch}{1.3}
\caption{Quantitative Results on DFC15 Multi-label Dataset (\%)}
\label{tab:dfc}
\centering
\begin{tabular}{ccccccc}
\Xhline{3\arrayrulewidth}
Model & m.$F_1$ & m.$F_2$ & m.$\text{P}_{e}$ & m.$\text{R}_{e}$ & m.$\text{P}_{l}$ & m.$\text{R}_{l}$\\
\hline
VGGNet \cite{simonyan2014very} & 73.86 & 74.09 & 76.16 & 74.95 & 62.57 & 59.95\\
VGGNet-RBFNN \cite{zeggada2017deep} & 72.21 & 73.02 & 74.08 & 74.42 & 60.82 & 66.58\\
CA-VGG-LSTM & 75.46 & 75.85 & 77.95 & 76.95 & 73.56 & 59.19\\
\textbf{CA-VGG-BiLSTM} & \textbf{76.25} & \textbf{76.93} & 78.27 & 78.30 & 74.99 & 64.31 \\
\hline
\hline
GoogLeNet \cite{szegedy2015going} & 74.99 & 73.41 & 81.01 & 73.01 & 71.80 & 53.95 \\
GoogLeNet-RBFNN \cite{zeggada2017deep}& 73.38 & 72.62 & 78.46 & 72.94 & 64.62 & 63.22\\
CA-GoogLeNet-LSTM & 75.67 & 75.46 & 79.08 & 76.12 & 70.22 & 60.65\\
\textbf{CA-GoogLeNet-BiLSTM} & \textbf{78.25} & \textbf{76.80} & 83.97 & 76.52 & 82.98 & 61.04\\
\hline
\hline
ResNet-50 \cite{he2016deep} & 78.10 & 76.21 & 84.89 & 75.64 & 81.50 & 59.99 \\
ResNet-RBFNN \cite{zeggada2017deep}& 78.36 & 78.08 & 82.64 & 78.76 & 72.01 & 69.85 \\
CA-ResNet-LSTM & 78.78 & 76.65 & 85.66 & 75.84 & 83.83 & 60.05 \\
\textbf{CA-ResNet-BiLSTM} & \textbf{83.65} & \textbf{80.61} & 91.93 & 79.12 & 94.35 & 62.35\\
\Xhline{3\arrayrulewidth}
\end{tabular}
\end{table}

Following the evaluation on UCM multi-label dataset, we assess our network on DFC15 multi-label dataset by calculating the mean $F_1$ and $F_2$ score as well as mean example- and label-based precision and recall. Table \ref{tab:dfc} shows experimental results on this dataset, and the conclusion can be drawn that modeling class dependencies with a bidirectional structure contributes significantly to multi-label classification. Specifically, the mean $F_1$ score achieved by CA-ResNet-BiLSTM is 4.87\% and 5.55\% higher than CA-ResNet-LSTM and ResNet, respectively. CA-VGG-BiLSTM obtains the best mean $F_1$ score of 76.25\% in comparison with VGGNet and CA-VGG-LSTM, and the mean $F_1$ score of CA-GoogLeNet-BiLSTM is 78.25\%, higher than its competitors. In comparison with Conv-RBFNN, CA-Conv-BiLSTM exhibits an improvement of at most 5.29\% and 4.18\% in terms of the mean $F_1$ and $F_2$ score, respectively. To conclude, all these increments demonstrate the effectiveness and robustness of our bidirectional structure for high-resolution aerial image multi-label classification. Several example predictions in DFC15 multi-label dataset are shown in Table \ref{tab:output}. The last two examples of DFC15 multi-label dataset show that trees are false negatives with the occurrence of vegetations due to their similar appearances. Moreover, we note the best result \cite{7536139} in 2015 IEEE GRSS Data Fusion Contest achieves 71.18\% in the mean F1 score, which is reduced by 12.47\% with respect to our best result. This is because predicting dense pixel-level labels is challenging in comparison with classifying multiple image-level labels.

\subsubsection{Qualitative Results}

\begin{figure*}[!t]
\captionsetup[subfigure]{labelformat=empty}
\centering
\subfloat{\includegraphics[width=0.102\textwidth]{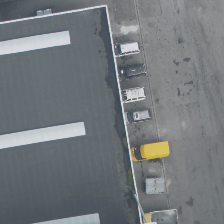}
\label{fig:vv1}}
\hspace{-0.4em}
\subfloat{\includegraphics[width=0.102\textwidth]{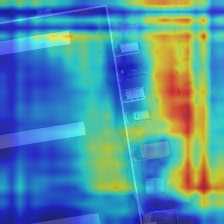}
\label{fig:vv2}}
\hspace{-0.4em}
\subfloat{\includegraphics[width=0.102\textwidth]{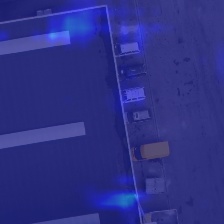}
\label{fig:vv3}}
\hspace{-0.4em}
\subfloat{\includegraphics[width=0.102\textwidth]{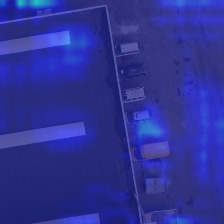}
\label{fig:vv4}}
\hspace{-0.4em}
\subfloat{\includegraphics[width=0.102\textwidth]{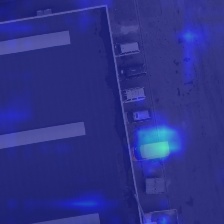}
\label{fig:vv5}}
\hspace{-0.4em}
\subfloat{\includegraphics[width=0.102\textwidth]{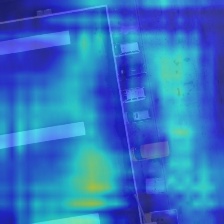}
\label{fig:vv6}}
\hspace{-0.4em}
\subfloat{\includegraphics[width=0.102\textwidth]{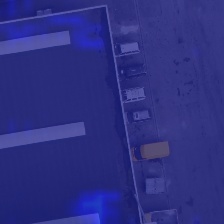}
\label{fig:vv7}}
\hspace{-0.4em}
\subfloat{\includegraphics[width=0.102\textwidth]{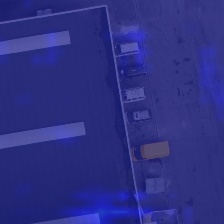}
\label{fig:vv8}}
\hspace{-0.4em}
\subfloat{\includegraphics[width=0.102\textwidth]{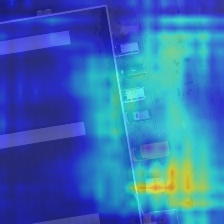}
\label{fig:vv9}}
\vspace{-0.5em}
\subfloat{\includegraphics[width=0.102\textwidth]{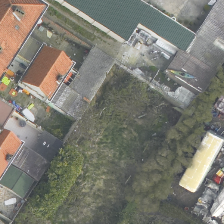}
\label{fig:vv10}}
\hspace{-0.4em}
\subfloat{\includegraphics[width=0.102\textwidth]{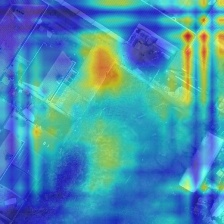}
\label{fig:vv11}}
\hspace{-0.4em}
\subfloat{\includegraphics[width=0.102\textwidth]{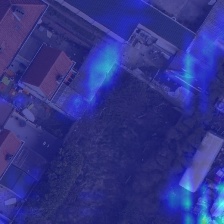}
\label{fig:vv12}}
\hspace{-0.4em}
\subfloat{\includegraphics[width=0.102\textwidth]{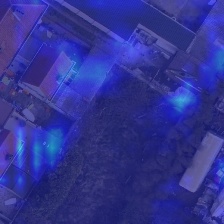}
\label{fig:vv13}}
\hspace{-0.4em}
\subfloat{\includegraphics[width=0.102\textwidth]{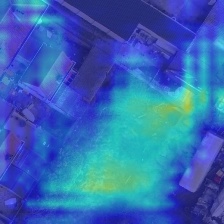}
\label{fig:vv14}}
\hspace{-0.4em}
\subfloat{\includegraphics[width=0.102\textwidth]{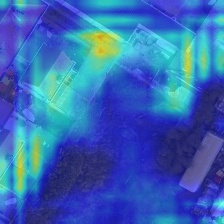}
\label{fig:vv15}}
\hspace{-0.4em}
\subfloat{\includegraphics[width=0.102\textwidth]{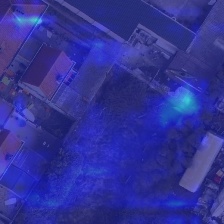}
\label{fig:vv16}}
\hspace{-0.4em}
\subfloat{\includegraphics[width=0.102\textwidth]{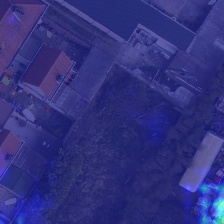}
\label{fig:vv17}}
\hspace{-0.4em}
\subfloat{\includegraphics[width=0.102\textwidth]{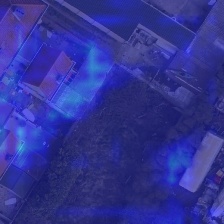}
\label{fig:vv18}}
\vspace{-0.5em}
\subfloat{\includegraphics[width=0.102\textwidth]{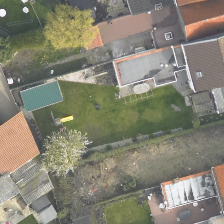}
\label{fig:vv19}}
\hspace{-0.4em}
\subfloat{\includegraphics[width=0.102\textwidth]{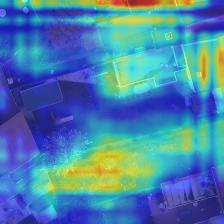}
\label{fig:vv20}}
\hspace{-0.4em}
\subfloat{\includegraphics[width=0.102\textwidth]{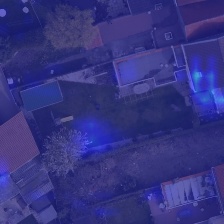}
\label{fig:vv21}}
\hspace{-0.4em}
\subfloat{\includegraphics[width=0.102\textwidth]{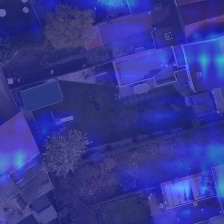}
\label{fig:vv22}}
\hspace{-0.4em}
\subfloat{\includegraphics[width=0.102\textwidth]{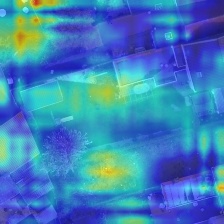}
\label{fig:vv23}}
\hspace{-0.4em}
\subfloat{\includegraphics[width=0.102\textwidth]{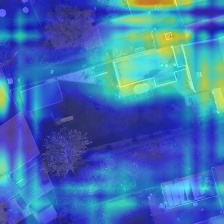}
\label{fig:vv24}}
\hspace{-0.4em}
\subfloat{\includegraphics[width=0.102\textwidth]{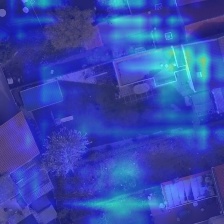}
\label{fig:vv25}}
\hspace{-0.4em}
\subfloat{\includegraphics[width=0.102\textwidth]{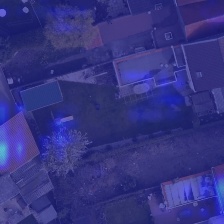}
\label{fig:vv26}}
\hspace{-0.4em}
\subfloat{\includegraphics[width=0.102\textwidth]{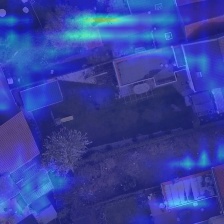}
\label{fig:vv27}}
\vspace{-0.5em}
\subfloat{\includegraphics[width=0.102\textwidth]{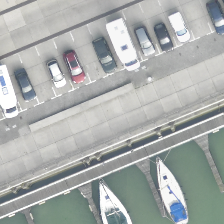}
\label{fig:vv28}}
\hspace{-0.4em}
\subfloat{\includegraphics[width=0.102\textwidth]{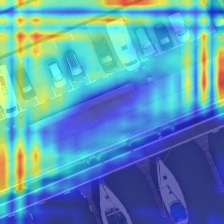}
\label{fig:vv29}}
\hspace{-0.4em}
\subfloat{\includegraphics[width=0.102\textwidth]{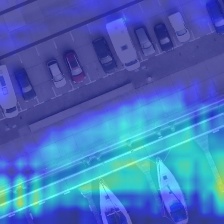}
\label{fig:vv30}}
\hspace{-0.4em}
\subfloat{\includegraphics[width=0.102\textwidth]{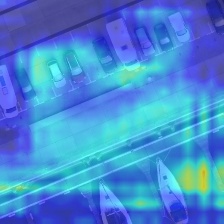}
\label{fig:vv31}}
\hspace{-0.4em}
\subfloat{\includegraphics[width=0.102\textwidth]{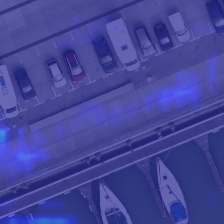}
\label{fig:vv32}}
\hspace{-0.4em}
\subfloat{\includegraphics[width=0.102\textwidth]{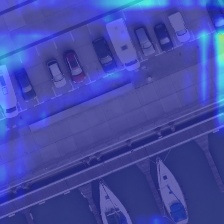}
\label{fig:vv33}}
\hspace{-0.4em}
\subfloat{\includegraphics[width=0.102\textwidth]{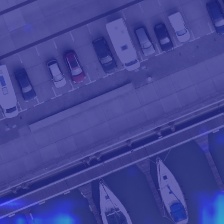}
\label{fig:vv34}}
\hspace{-0.4em}
\subfloat{\includegraphics[width=0.102\textwidth]{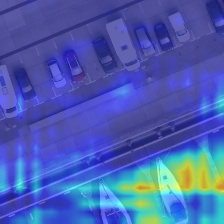}
\label{fig:vv35}}
\hspace{-0.4em}
\subfloat{\includegraphics[width=0.102\textwidth]{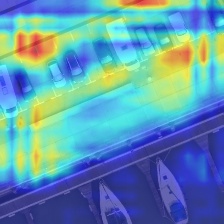}
\label{fig:vv36}}
\vspace{-0.5em}
\subfloat[(a)]{\includegraphics[width=0.102\textwidth]{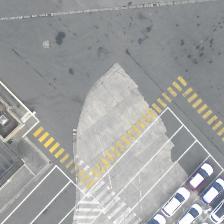}
\label{fig:vv37}}
\hspace{-0.4em}
\subfloat[(b)]{\includegraphics[width=0.102\textwidth]{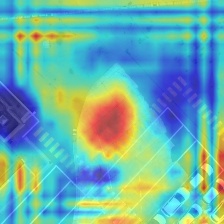}
\label{fig:vv38}}
\hspace{-0.4em}
\subfloat[(c)]{\includegraphics[width=0.102\textwidth]{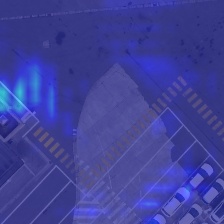}
\label{fig:vv39}}
\hspace{-0.4em}
\subfloat[(d)]{\includegraphics[width=0.102\textwidth]{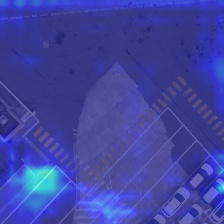}
\label{fig:vv40}}
\hspace{-0.4em}
\subfloat[(e)]{\includegraphics[width=0.102\textwidth]{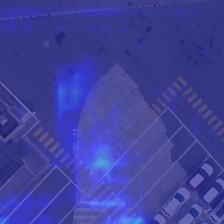}
\label{fig:vv41}}
\hspace{-0.4em}
\subfloat[(f)]{\includegraphics[width=0.102\textwidth]{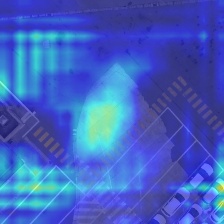}
\label{fig:vv42}}
\hspace{-0.4em}
\subfloat[(g)]{\includegraphics[width=0.102\textwidth]{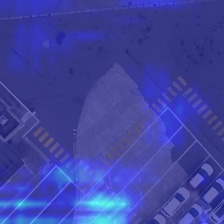}
\label{fig:vv43}}
\hspace{-0.4em}
\subfloat[(h)]{\includegraphics[width=0.102\textwidth]{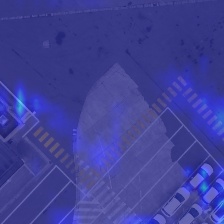}
\label{fig:vv44}}
\hspace{-0.4em}
\subfloat[(i)]{\includegraphics[width=0.102\textwidth]{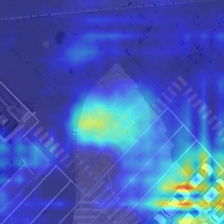}
\label{fig:vv45}}
\caption{Example class attention maps of (a) images in DFC15 dataset with respect to (b) impervious, (c) water, (d) clutter, (e) vegetation, (f) building, (g) tree, (h) boat, and (i) car. Red indicates strong activations, while blue represents non-activations. Besides, normalization is performed based on each row for a fair comparison among class attention maps of the same images.}
\label{fig:visual2}
\end{figure*}

To study the effectiveness of class-specific features, we visualize class attention maps learned from the proposed class attention learning layer, as shown in Fig. \ref{fig:visual2}. Columns (b)-(i) are example class attention maps with respect to (b) impervious, (c) water, (d) clutter, (e) vegetation, (f) building, (g) tree, (h) boat, and (i) car. As we can see, figures at column (b) of Fig. \ref{fig:visual2} show that the network pays high attention to impervious regions, such as parking lots, while figures at column (i) highlight regions of cars. However, some of class attention maps for negative object labels exhibit unexpected strong activations. For instance, the class attention map for the car at the third row of Fig. \ref{fig:visual2} is not supposed to highlight any region due to its absence of cars. This can be explained as the highlighted regions share similar patterns as cars, which also illustrates why the network made wrong predictions (cf. wrongly predicted car label in Fig. \ref{fig:visual2}). Overall, the visualization of class attention maps demonstrates that the features captured from the proposed class attention learning layer are discriminative and class-specific. Besides, we note that there exist strong border artifacts in figures, especially those at column (b) of Fig. \ref{fig:visual2}, which questions whether improving the quality of class attention maps benefits the effectiveness of the BiLSTM-based sub-network. Then we experimented with using the skip connection scheme in order to refine class attention maps. Experimental results demonstrated that this provides negligible improvements.

\section{Conclusion}
\label{sec:conclusion}
In this paper, we propose a novel network, CA-Conv-BiLSTM, for the multi-label classification of high-resolution aerial imagery. The proposed network is composed of three indispensable elements: 1) a feature extraction module, 2) a class attention learning layer, and 3) a bidirectional LSTM-based sub-network. Specifically, the feature extraction module is responsible for capturing fine-grained high-level feature maps from raw images, while the class attention learning layer is designed for extracting discriminative class-specific features. Afterwards, the bidirectional LSTM-based sub-network is used to model the underlying class dependency in both directions and predict multiple object labels in a structured manner. With such design, the prediction of multiple object-level labels is performed in an ordered procedure, and outputs are structured sequences instead of discrete values. We evaluate our network on two datasets, UCM multi-label dataset and DFC15 multi-label dataset, and experimental results validate the effectiveness of our model from both quantitative and qualitative respects. On one hand, the mean $F_2$ score is increased by at most 0.0446 compared to other competitors. On the other hand, visualized class attention maps, where discriminative regions for existing objects are strongly activated, demonstrate that features learned from this layer are class-specific and discriminative. Looking into the future, the application of our network can be extended to fields, such as weakly supervised semantic segmentation and object localization.

\section*{Acknowledgment}

This work is jointly supported by the China Scholarship Council, the Helmholtz Association under the framework of the Young Investigators Group SiPEO (VH-NG-1018, www.sipeo.bgu.tum.de), and the European Research Council (ERC) under the European Union's Horizon 2020 research and innovation programme (grant agreement No. ERC-2016-StG-714087, Acronym: \textit{So2Sat}). In addition, the authors would like to thank the National Center for Airborne Laser Mapping and the Hyperspectral Image Analysis Laboratory at the University of Houston for acquiring and providing the data used in this study, and the IEEE GRSS Image Analysis and Data Fusion Technical Committee.


\section*{References}
\bibliographystyle{elsarticle-num}
\bibliography{opticalSAR.bib}

\end{document}